\documentclass[11pt,3p,review,authoryear]{elsarticle}

\usepackage{amssymb}
\usepackage{amsmath}
\usepackage{dsfont}
\usepackage{booktabs}
\usepackage{adjustbox}
\usepackage{ragged2e} 
\usepackage{xcolor}
\usepackage{float}
\usepackage{pdflscape}
\usepackage{caption}
\usepackage{subcaption}
\usepackage{array}
\usepackage{svg}
\usepackage{cleveref}
\usepackage{csquotes} 

\usepackage{amssymb}
\usepackage{amsmath}
\usepackage{dsfont}
\usepackage{booktabs}
\usepackage{adjustbox}
\usepackage{ragged2e} 
\usepackage{xcolor}
\usepackage{float}
\usepackage{pdflscape}
\usepackage{array}
\usepackage{tikz}
\usepackage[pagewise]{lineno} 
\usepackage{csquotes}

\journal{International Journal of Forecasting}

\makeatletter
\def\ps@pprintTitle{%
  \let\@oddhead\@empty
  \let\@evenhead\@empty
  \let\@oddfoot\@empty
  \let\@evenfoot\@oddfoot}
\makeatother

\begin{document}

\begin{frontmatter}

\title{Automobile demand forecasting: Spatiotemporal and
hierarchical modeling, life cycle dynamics, and
user-generated online information}


\address[ss]{School of Management, Technical University of Munich, Arcisstraße
21, Munich, 80333, Bavaria, Germany}
\address[bmw]{BMW AG, Petuelring 130, Munich, 80809, Bavaria, Germany}
\address[bb]{Munich Data Science Institute, Walther-von-Dyck-Straße
10, Garching, 85748, Bavaria, Germany}

\author[ss,bmw]{Tom Nahrendorf\corref{cor}}
\ead{tom.nahrendorf@tum.de}
\author[ss,bb]{Stefan Minner}
\author[bmw]{Helfried Binder}
\author[bmw]{Richard Zinck}


\begin{abstract}
Premium automotive manufacturers face increasingly complex forecasting challenges due to high product variety, sparse variant-level data, and volatile market dynamics. This study addresses monthly automobile demand forecasting across a multi-product, multi-market, and multi-level hierarchy using data from a German premium manufacturer. 
The methodology combines point and probabilistic forecasts across strategic and operational planning levels, leveraging ensembles of LightGBM models with pooled training sets, quantile regression, and a mixed-integer linear programming reconciliation approach. Results highlight that spatiotemporal dependencies, as well as rounding bias, significantly affect forecast accuracy, underscoring the importance of integer forecasts for operational feasibility. Shapley analysis shows that short-term demand is reactive, shaped by life cycle maturity, autoregressive momentum, and operational signals, whereas medium-term demand reflects anticipatory drivers such as online engagement, planning targets, and competitive indicators, with online behavioral data considerably improving accuracy at disaggregated levels.
\end{abstract}


\end{frontmatter}

\section{Introduction}

Premium original equipment manufacturers (OEMs) pursue mass customization strategies, offering substantially greater product variety than volume manufacturers. On average, German premium OEMs provide more than five times as many factory-fitted options as their Japanese volume-brand counterparts \citep{Staeblein2015}. Such strategies lead to increasingly diverse product portfolios, resulting in greater demand fragmentation, increasing uncertainty at the individual product level, and higher costs associated with supply--demand mismatches \citep{Moreno2016}. Consequently, the forecasting problem for premium OEMs extends beyond aggregate product-level demand to a large number of low-volume variants, referred to as product types hereafter. The combination of high dimensionality and sparse historical data for many configurations further amplifies forecast uncertainty and poses substantial challenges for supply chain planning.

Automobile demand forecasts play a central role in supporting both strategic and operational decision-making. Key applications include production and capacity planning, pricing, inventory management, and marketing. These decisions span different organizational levels and time horizons, requiring forecasts across multiple levels of the product and market hierarchy, from aggregated brands and regions to specific product types. While high-level forecasts inform long-term strategic planning, lower-level forecasts are critical for short- to medium-term operational decisions. The value of improved forecasting lies not only in higher predictive accuracy, but also in more effective decision-making through reduced operational costs, more efficient resource allocation, and better risk management. In particular, probabilistic forecasts are essential for risk-aware decision-making, as they quantify forecast uncertainty, which is crucial for applications such as pricing, capacity buffering, and inventory control. Quantile regression has thus gained attention as a means of modeling the full distribution of demand, rather than relying solely on point forecasts.

Despite its importance, automotive demand forecasting presents several distinctive and underexplored challenges. First, strict product hierarchies require coherent forecasts across multiple levels. Forecasting models should explicitly ensure this structural consistency. Second, demand patterns are shaped by complex life cycle dynamics: five- to seven-year product cycles, mid-cycle facelifts, and overlapping launches generate irregular, multi-peak demand curves that are difficult to capture with standard life cycle models. Third, spatiotemporal correlations across geographic regions and over time significantly influence demand, yet remain difficult to model effectively. Finally, the increasing digitalization of the automotive sales process enables demand sensing through user-generated data, such as online configurator website traffic, although their predictive value remains unclear in the forecasting literature \citep{Schaer2019}.

While hierarchical forecasting, life cycle dynamics, spatiotemporal dependencies, and behavioral data have each been studied independently, they have not been jointly addressed within a real-world automotive forecasting framework. To fill this gap, we present an empirical study conducted in collaboration with a German premium OEM and investigate the following research questions: (RQ1) Which point and probabilistic forecasting methods yield the highest accuracy in predicting automobile demand at various hierarchical levels? (RQ2) What are the key spatiotemporal, endogenous, and exogenous factors influencing automobile demand? (RQ3) To what extent does user-generated online information improve forecast accuracy?

To answer these questions, we apply light gradient boosting machine learning (LightGBM; \citealp{Ke2017}) models with quantile regression to estimate the full distribution of demand, extending the global architecture used in the M5 competitions \citep{Makridakis2022,Makridakis2022a}. Building on this foundation, we introduce a probabilistic pooled ensemble (direct-recursive averaging) and replace fixed hierarchical pooling with data-driven pooling-set selection that trades off accuracy and complexity. To address the variant-level forecasting challenges of premium OEMs, we propose a mixed-integer linear programming (MILP) integer-coherent reconciliation method. This ensures coherent forecasts across the hierarchy while preserving integer demand values, avoiding rounding errors that can distort manufacturing and supply chain operations. We further employ explainable artificial intelligence (XAI) to identify influential demand drivers (RQ2) and evaluate the contribution of user-generated online information through statistical testing (RQ3).

The remainder of the paper is structured as follows: Section 2 reviews related work and theoretical foundations. Section 3 presents the modeling, forecasting, and data pooling strategies employed, along with our proposed reconciliation approach. Section 4 describes the empirical case study, including the data characteristics and the forecast evaluation metrics. Section 5 reports the results addressing RQ1--RQ3. Finally, Section 6 concludes the paper and outlines directions for future research.

\section{Related literature}
\subsection{Automotive demand forecasting and probabilistic methods}
Prior research in the automotive domain has predominantly addressed macroeconomic modeling of total vehicle sales \citep[e.g.,][]{Berkovec1985}, using dynamic simulation frameworks to study aggregate demand patterns. More recent work has targeted brand-level \citep[e.g.,][]{Fantazzini2015} or component-level forecasts, such as spare parts and batteries \citep[e.g.,][]{Bottani2021, Goncalves2021}, employing machine learning (ML) approaches. These studies have relied on aggregated data and point forecasts. However, high-granularity demand forecasting at the market-product type level remains underexplored, particularly for premium OEMs and in settings that require quantification of forecast uncertainty.

Quantile regression methods have gained traction for their ability to estimate full conditional distributions rather than single-point estimates. Recent studies demonstrate the effectiveness of tree-based quantile models across various domains: \citet{Arora2023} applied quantile regression forests (QRF) to predict patient waiting times in emergency departments, which led to a more uniform patient load across the network. \citet{Guo2021} used regression trees to forecast airport transfer times, leading to reduced staffing costs, while \citet{Salari2022} showed that cost-sensitive QRFs improved customer satisfaction and boosted e-commerce sales through better delivery time prediction. The value of quantile-based forecasting has also been confirmed in recent forecasting competitions. In the M5 Uncertainty competition, the winning submission \citep{Lainder2022} employed gradient-boosted decision trees to generate probabilistic forecasts across a large retail hierarchy. In the related M5 Accuracy competition, the Direct Recursive Forecast Averaging via Partial Pooling model (DRFAM-PP; \citealp{In2022}) delivered strong point-forecast performance by combining direct and recursive strategies with pooled training datasets. 

We extend DRFAM-PP from point to quantile forecasting, training direct and recursive learners as quantile models and averaging forecasts to obtain predictive distributions, thereby quantifying forecast uncertainty for downstream decision making. We replace fixed pooling with data-driven pooling set selection, casting pooling as a binary partitioning problem that endogenously opens pools from a candidate family using validation losses and a complexity-aware cost, allowing us to identify heterogeneous pools. We validate the approach in a real-world case at a German premium OEM, delivering probabilistic forecasts at the market–product-type level, a setting not yet addressed in the literature, and show that purely hierarchical pooling is empirically sub-optimal.

\subsection{Spatiotemporal modeling}
Traditional statistical time series regression models are adept at handling simple historical demand observations. However, they often struggle to capture the complexities inherent in multi-dimensional spatiotemporal correlated data \citep{Liang2018}. If OEMs can obtain prior information regarding the demand distribution across various spatial areas and time intervals, they can better align supply with market demands.

The literature primarily focuses on spatiotemporal characteristics within transportation applications, employing ML approaches for modeling. Notable applications include transportation demand forecasting in car-hailing \citep[][]{Guo2022}, bike-sharing \citep[][]{Yao2018, Gammelli2022}, taxi traffic flow prediction \citep{Ding2023}, and ridesplitting \citep{Li2024}. For instance, \cite{Guo2022} proposed a forecasting approach that integrates spatial, temporal, and meteorological features. Experiments conducted on real-world data reveal that random forest regression, support vector regression, and k-nearest neighbor regression achieve the highest forecast accuracy.
\cite{Li2024} employed multi-graph convolutional neural networks to extract spatiotemporal correlations among ride orders, demonstrating superior performance over single-graph architectures. \cite{Bi2022} proposed a retail sales forecasting approach at the store-product level, introducing a temporal latent factor model that achieves accurate predictions through a single tensor factorization across multiple stores and products.

While spatiotemporal modeling has been widely applied in urban transportation and retail contexts, existing research has primarily focused on local proximities (e.g., cities or stores within regions). In contrast, real-world industrial problems in the automotive sector involve multi-market and multi-product structures, where correlations arise not only from geography but also from shared economic conditions, product availability, and synchronized marketing strategies. Such cross-market interactions remain seldom addressed in the forecasting literature. This study extends spatiotemporal modeling to settings where markets are economically or behaviorally aligned rather than geographically adjacent, thereby moving beyond traditional locality-based assumptions.

\subsection{Hierarchical forecasting}
Base forecasts in hierarchical time series can be generated using local or global models. Local models capture series-specific patterns but become computationally intensive as the hierarchy expands, whereas global models exploit cross-series information and generally achieve higher accuracy and scalability \citep{MonteroManso2021}.

Once base forecasts are generated, reconciliation methods are used to ensure coherence across the hierarchy. Comprehensive reviews of forecast reconciliation methods can be found in \cite{Athanasopoulos2024}, \cite{Hollyman2021}, and specifically for demand forecast reconciliation, \cite{Babai2022}. Traditional single-level reconciliation approaches such as bottom-up (BU), top-down (TD), and middle-out (MO) reconcile forecasts by either aggregating or disaggregating series. However, these methods rely on limited information from a specific hierarchical level and have shown limited robustness over time \citep{Anderer2022}. 

To address the limitations of single-level methods, least squares and optimization-based reconciliation techniques have gained popularity. \cite{Hyndman2011} forecast each level independently and then use ordinary least squares (OLS) to optimally combine the forecasts, which was later extended to a weighted least squares (WLS) solution \citep{Hyndman2016} that weights series by the variance of the base forecast. 
\cite{Wickramasuriya2019} introduced an optimization strategy that minimizes the trace (MinT) of the covariance matrix of forecast errors, encompassing both OLS and WLS as special cases. While these methods assume unbiased base forecasts, \cite{BenTaieb2019} proposed an empirical risk minimization approach that minimizes the average squared error between reconciled forecasts and actual values, achieving competitive accuracy relative to benchmark methods under the unbiasedness assumption. However, traditional methods like OLS, WLS, and MinT do not guarantee non-negativity of reconciled forecasts, which can be addressed through constrained quadratic programming  \cite[e.g.,][]{Wickramasuriya2020}. 

Most reconciliation methods yield continuous forecasts, which are unsuitable for planning contexts that require integer decisions such as material requirements planning, car sequencing, and dealer order promising. This is critical for premium OEMs, where demand at the market–product type level involves low volumes and high variance. Rounding fractional reconciled forecasts both breaks hierarchical coherence and distorts planning, since even a ±1 unit error can misrepresent a substantial share of demand and lead to costly buffers in parts inventory and production capacity. To address this, we propose a MILP formulation for discrete forecast reconciliation that is less complex, computationally efficient compared to existing approaches \citep[e.g.,][]{Zhang2024}, and yields forecasts that are both hierarchical consistent and operationally feasible. Our approach is decision-focused and follows empirical risk minimization in the hierarchical setting \citep{BenTaieb2019}: it minimizes an out-of-sample average weighted $L_1$ loss with series-reliability and level-importance weights, thereby protecting stable aggregates while shifting adjustments to noisier components. Geometrically, it is the weighted-$L_1$ analogue of MinT on the integer coherent set; its LP relaxation recovers continuous $L_1$ reconciliation on the coherent cone, and under a quadratic loss with base-error covariance it connects to MinT, reducing to OLS when the covariance is proportional to the identity matrix. Beyond integrality, the formulation can be extended to incorporate domain-specific linear constraints (e.g., forecast immutability), making it applicable across domains and allowing reconciliation and feasibility to be addressed within a single optimization.

\subsection{Life cycle dynamics}
The dynamic nature of product life cycles introduces significant complexity to the forecasting process \citep[with automotive manufacturing as a documented example;][]{Georgiadis2006}. 

While early model-based approaches used differential equations to describe product adoption \citep[e.g.,][]{Kurawarwala1996, Lotfi2022}, recent studies have shifted toward data-driven methods. \cite{Hu2018} fit product life cycle (PLC) curves to historical data, clustering similar patterns and identifying fourth-order polynomials as the best fit. Similarly, \cite{Elalem2023} employ data-driven time series clustering and explored ML approaches for PLC fitting. Their results suggest that while autoregressive integrated moving average models with exogenous inputs (ARIMAX) outperform deep neural networks in terms of mean absolute errors, the latter demonstrate robust performance under noisy data conditions. In contrast, \cite{Goncalves2021} report that ARIMAX models provide higher accuracy during the early stages of the PLC, whereas ML models excel in later stages, indicating that deep learning is not uniformly superior. \cite{Guo2024} propose an ensemble that combines Bayesian model averaging with an exponential-smoothing life-cycle component, which adapts to recent changes and outperforms ML and Bayesian updating methods in forecasting the full demand distribution of new products.

Across industries, product life cycles occur as items are introduced, updated, and ultimately withdrawn (e.g., electronics, pharmaceuticals, software, media, manufacturing). Prior studies \citep[e.g.,][]{Guo2024,Hu2018,Lotfi2022} often impose parametric PLC shapes (e.g., polynomial functions, Bass diffusion) and cluster pre-shaped trajectories, thereby limiting adaptability to the heterogeneous, multi-peak patterns driven by product updates and market shocks. Moreover, the studies reviewed operate at a single level, with PLC features defined per item, which restricts higher-level models from exploiting fine-grained temporal dynamics observed at lower levels. In contrast, we adopt a shape-agnostic, domain-neutral representation: the age--volume moment (AVM), defined as the product of time-since-introduction and realized volume. AVM requires only an introduction date and an observable measure of volume (e.g., units, downloads, prescriptions, streams) and is additive across the hierarchy, enabling consistent learning across items, product lines, and markets without imposing a predefined PLC curve. 

\subsection{User-generated online information}
In the automotive industry, user-generated online information, such as website traffic and online search trends, has gained increasing attention for demand forecasting, as manufacturers and dealers seek to better capture consumer preferences and behaviors. However, the existing literature provides mixed evidence on its effectiveness. For instance, \cite{Du2015} utilized online search trends to forecast car demand in the U.S. automotive market, demonstrating substantial improvements
in model fit. Similarly, \cite{Fantazzini2015} developed multivariate models that integrate economic indicators and Google Search data to predict monthly car sales for several German OEMs. Their results indicate that models incorporating online information consistently outperform benchmark models that do not include such data across various car brands and forecast horizons. \cite{Zhang2022} used online reviews and search engine data to forecast aggregated product-level automobile sales, converting reviews into sentiment scores and training a backpropagation neural network, which significantly improved forecast accuracy and robustness. 

In contrast, \cite{Schaer2019} critically assess studies such as \cite{Du2015} and \cite{Fantazzini2015}, highlighting key design weaknesses (e.g., the absence of adequate benchmark models). They caution against overestimating the value of online information for operational decision-making. Notably, their findings show that in some cases, simple univariate time series models can outperform more complex models that incorporate online data, raising important questions about the robustness and practical value of these behavioral information sources.

No prior research has used online car configurator data to predict automotive demand, leaving a clear gap in the literature. We address this gap by introducing configurator interactions as a novel source of user-generated information for demand forecasting. Our contribution is twofold: first, we statistically evaluate the predictive value of configurator data and demonstrate its ability to enhance forecast accuracy when integrated into existing models; second, we apply XAI techniques to interpret how configurator usage patterns relate to future demand, providing transparent and actionable insights for decision-makers.

\section{Automobile demand forecasting}\label{chap: methods}
\subsection{Modeling}
We formulate the task of predicting automobile demand as a multivariate, multi-step time series forecasting problem. Let $\{\mathbf{y}_t\}_{t=1}^T = \{\mathbf{y}_1, \mathbf{y}_2, \dots, \mathbf{y}_T\}$ represent the automobile demand time series (dependent variable) and $\{\mathbf{x}_t\}_{t=1}^T = \{\mathbf{x}_1, \mathbf{x}_2, \dots, \mathbf{x}_T\}$ denote the exogenous series (independent variable) over the past $T$ time periods. Here, $\mathbf{y}_t = \{y_t^1, y_t^2, \dots, y_t^n\} \in \mathbb{R}^n$ defines a vector of demand values for $n$ different time series in period $t$. Similarly, $\mathbf{x}_t = \{\mathbf{x}_t^1, \mathbf{x}_t^2, \dots, \mathbf{x}_t^n\} \in \mathbb{R}^{n \times m}$ defines a vector of $m$ exogenous variables (features, covariates, independent variables) in period $t$. We approximate the unknown relationship between the dependent variable 
and its covariates by developing a forecasting model $f(\cdot)$, trained on the dataset $\mathcal{D} = \{(\mathbf{y}_t, \mathbf{x}_t) \}_{t=1}^T$. The objective is to minimize the difference between the observed values and the model predictions:

 \begin{equation}
   \big\{\mathbf{\hat{y}}_t\big\}_{t=T+1}^{T+H} = \{\mathbf{\hat{y}}_{T+1}, \mathbf{\hat{y}}_{T+2}, \dots, \mathbf{\hat{y}}_{T+H}\} = f\Big(\{\mathbf{y}_t\}_{t=1}^T, \{\mathbf{x}_t\}_{t=1}^T; \theta \Big),
\end{equation}
where $H$ is the forecast horizon and $\theta$ is a parameter vector of the model $f(\cdot)$.

While point forecasts provide a single expected value for future demand, they do not capture the uncertainty inherent in real-world forecasting tasks. To address this limitation, quantile forecasting enables the estimation of the distribution of automobile demand. We define the predicted automobile demand $\{\mathbf{\hat{y}}_t^q\}_{t=T+1}^{T+H} = \{\mathbf{\hat{y}}_{T+1}^q,\mathbf{\hat{y}}_{T+2}^q, \dots, \mathbf{\hat{y}}_{T+H}^q\}$ for a given quantile $q \in (0,1)$ as follows:

\begin{equation}
   \big\{\mathbf{\hat{y}}_t^q\big\}_{t=T+1}^{T+H} = \Big\{\inf \big(\mathbf{y}_t \in \mathbb{R}^n: q \leq \hat{F}(\mathbf{y}_t |  \mathbf{y}_1, \dots, \mathbf{y}_{t-1}, \mathbf{x}_t)\big) \Big\}_{t=T+1}^{T+H},
\end{equation}

where $\hat{F}(\cdot)$ denotes the estimated cumulative distribution function of the conditional distribution $\mathbb{P}\big(\mathbf{y}_t | \mathbf{y}_1, \dots, \mathbf{y}_{t-1}, \mathbf{x}_t \big)$ in period $t$. The distribution $\mathbb{P}(\cdot)$ is conditioned on the autoregressive parts $\mathbf{y}_1, \dots, \mathbf{y}_{t-1}$ and exogenous variables $\mathbf{x}_t$. Thus, we can investigate the influence of explanatory variables on the shape of the demand distribution. To provide a comprehensive view of the distribution, we focus on the median and prediction intervals (PIs) at the 50\%, 67\%, 95\%, and 99\% levels by investigating a set of quantiles $\mathcal{Q} = \{0.005,0.025,0.165,0.25,0.5,0.75,0.835,0.975,0.995\}$. The median, along with the 50\% and 67\% PIs, capture the central tendency and short-term uncertainty of automobile demand. This information is useful for monthly vehicle allocation decisions, where manufacturers must balance limited supply against expected regional demand to minimize inventory costs and lost sales. The 95\% and 99\% PIs provide insight into the extreme tails of the distribution. Their upper and lower bounds are crucial for risk assessment and capacity planning, enabling stakeholders to prepare for rare but impactful demand fluctuations. 

\subsection{Pooling strategies}
Pooling combines information across related time series to improve forecast accuracy. In practice, this spans local models (no pooling), global models (full pooling), and intermediate partial pooling along structured groups. Recent work \citep{MonteroManso2021} reframes pooled forecasting as a partitioning problem on the local–global spectrum, where the granularity of the partition governs model complexity and generalization and should be learned from data rather than fixed a priori.

We therefore formulate pooling as a data-driven partitioning problem: series are assigned to pools that share one forecast model, trading off empirical accuracy and model complexity. Let $\mathcal{I}$ be the set of series and $\mathcal{G}$ a finite family of candidate pools, with $g \subseteq \mathcal{I}$. The candidate family integrates hierarchy-derived groups (e.g., market–product cluster), domain categories (e.g., fuel and body type), data-driven clusters, and their combinations. To guarantee coverage and a no-pooling fallback, we include all singletons $\{ i \}$ (local models), so that every $i \in \mathcal{I}$ belongs to at least one $g \in \mathcal{G}$. For each admissible pair $(i,g)$ with $i \in g$, we compute a leakage-safe rolling-origin cross-validated loss $L_{ig}$ on validation folds only. Each pool $g$ incurs an opening cost $\kappa_g$ that penalizes (i) the number of models induced by the grouping (computational complexity) and (ii) within-group sample imbalance (overfitting risk). 

We define the pooling selection as a binary partitioning problem (POOL-SEL-BP):
\begin{equation}\label{eqn:obj}
    \min_{x \in \{0,1\}^{\mathcal{I} \times \mathcal{G}}, y \in \{0,1\}^{\mathcal{G}}} \sum_{i \in \mathcal{I}}\sum_{g \in \mathcal{G}: i \in g} L_{ig} x_{ig} + \lambda \sum_{g \in \mathcal{G}} \kappa_g y_g,
\end{equation}
s.t.
\begin{equation}\label{eqn: pp const1}
     \sum_{g \in \mathcal{G}: i \in g} x_{ig} = 1, \quad \forall i\in \mathcal{I},
\end{equation}
\begin{equation}\label{eqn: pp const2}
     x_{ig} \leq y_g, \quad \forall i \in \mathcal{I}, g \in \mathcal{G},
\end{equation}

with $\lambda\geq0$ trading off accuracy and complexity. The selected pooling families and the exact pool-cost specifications are provided in the e-companion (EC.3).

\subsection{Forecasting strategies and ensemble construction}
We consider three multi-step forecasting strategies, denoted as $\mathcal{S} = \{\text{DIR}, \text{REC}, \text{HYB}\}$. Each generates forecasts at horizon $h \in [1,H]$, using only information available up to $T$. All strategies are trained separately for each quantile $q \in \mathcal{Q}$, enabling distributional forecasts without altering the underlying model structure.

Direct forecasting (DIR) trains a separate model for each forecast horizon:
\begin{equation}
    \mathbf{\hat{y}}_{T+h}^{\text{DIR}} = f_h\big(
    \mathbf{y}_{1:T}, \mathbf{x}_{T}; \theta_h
    \big), 
\end{equation}
where $f_h(\cdot)$ is a horizon-specific model with parameters $\theta_h$. This strategy avoids error accumulation but does not enforce temporal consistency across horizons \citep{BenTaieb2012}. Recursive forecasting (REC) trains a single one-step-ahead model and recursively uses previous forecasts as inputs:
\begin{equation}
    \mathbf{\hat{y}}_{T+h}^{\text{REC}} = f\big(
    \mathbf{\overline{y}}_{T+h-1}, \mathbf{\overline{x}}_{T+h-1}; \theta
    \big), 
\end{equation}
where $\overline{\mathbf{y}}_{T+h-1}= \{\mathbf{y_1}, \dots, \mathbf{y_T}, \mathbf{\hat{y}}_{T+1}^{\text{REC}}, \dots, \mathbf{\hat{y}}_{T+h-1}^{\text{REC}}\}$ denotes the sequence of observed and predicted targets, and $\overline{\mathbf{x}}_{T+h-1}$ are the corresponding exogenous features, replacing unknown values with forecasts when needed. While compact and efficient, REC suffers from error accumulation \citep{BenTaieb2012}. Hybrid forecasting (HYB) combines the strengths of DIR and REC by training a separate model for each horizon while incorporating lagged forecasts as autoregressive inputs:
\begin{equation}
    \mathbf{\hat{y}}_{T+h}^{\text{HYB}} = f_h\big(
    \mathbf{\overline{y}}_{T+h-1}, \mathbf{x}_{T}; \theta_h
    \big). 
\end{equation} 
HYB (often referred to as \textit{DirRec}) combines the horizon-specific modeling flexibility of DIR with the temporal dependency modeling of REC. As shown in \citet{BenTaieb2012}, such hybrid models can reduce cumulative forecast errors while preserving temporal structures. To further improve prediction accuracy and robustness, we define an ensemble model (ENS) that averages forecasts from all strategies in $\mathcal{S}$:
\begin{equation}
\mathbf{\hat{y}}_{T+h}^{\text{ENS}} = \frac{1}{|\mathcal{S}|} \sum_{s \in \mathcal{S}} \mathbf{\hat{y}}_{T+h}^s
\end{equation}
By combining forecasts from diverse models, ENS leverages their complementary strengths to reduce variance and mitigate individual model biases \citep{Breiman1996}. 

Model diversity is key to ensemble effectiveness \citep{Brown2005}. To further enhance this, we include a pooled ensemble based on hierarchical pooling (DRFAM-PP (HIER)) and optimal pooling selection via POOL-SEL-BP (DRFAM-PP (OPT)). We define the DRFAM-PP forecast for series $i$ at horizon $h$ as the average of direct and recursive forecasts across all active pools that include series $i$:

\begin{equation}
\mathbf{\hat{y}}_{i,T+h}^{\text{DRFAM-PP}} = \frac{1}{|\mathcal{S}_{\text{DR}}||\mathcal{G}_i|} \sum_{s \in \mathcal{S}_{\text{DR}}} \sum_{g \in \mathcal{G}_i} \mathbf{\hat{y}}_{T+h}^{s, g},
\end{equation}

where $\mathcal{S}_{\text{DR}} = \{\text{DIR}, \text{REC}\}$ denotes the set of included forecasting strategies, $\mathcal{G}_i = \{ g \in \mathcal{G} | i \in g, y_g =1 \}$ is the set of active pools containing series $i$, and $\hat{\mathbf{y}}_{T+h}^{s,g}$ is the forecast at horizon $h$ from strategy $s$ trained on the pooled dataset $\mathcal{D}_g$. HYB models are excluded from DRFAM-PP to remain consistent with its original definition. However, we evaluate them separately and also include them in broader ensemble benchmarks.

\subsection{Forecast reconciliation}
Forecast reconciliation ensures that disaggregated automobile demand forecasts (e.g., market–product type) are coherent with their aggregates (brand, market). Following \cite{Hyndman2011}, reconciled forecasts can be written as $\tilde{\mathbf{y}}_t = \mathbf{S}\mathbf{G}\hat{\mathbf{y}}_t$. Here, $\tilde{\mathbf{y}}_t$ denotes the reconciled vector at horizon $t$; $\mathbf{S}\in \{0,1\}^{N \times B}$ is the structural/summing matrix that maps the $B$ bottom-level series to all $N$ series (so the coherent subspace is $s=\mathrm{col}(\mathbf{S})$); and matrix $\mathbf{G} \in \mathbb{R}^{B \times N}$ defines the reconciliation weights and
determines how base forecasts $\mathbf{\hat{y}}$ are adjusted to ensure coherence across the hierarchy. Classical methods (e.g., MinT) can then be interpreted as weighted $L_2$ projections of the base forecasts onto the coherent subspace $s$, and, when non-negativity is imposed, onto the coherent cone $K = s\cap \mathbb{R}^N_{\ge 0}$ \citep{Panagiotelis2021}. We adopt the same geometric lens, but change both the metric and the support: we use a weighted $L_1$ distance and restrict the solution to the integer lattice inside the cone $K_{\mathbb{Z}} = K \cap \mathbb{Z}^N = (s \cap \mathbb{R}_{\geq 0}) \cap \mathbb{Z}^N$. The geometric motivation and a side-by-side comparison with OLS and MinT are further illustrated in the e-companion EC.4.

In the automotive industry, forecasts must be non-negative and integer-valued, reflecting the discrete nature of vehicle production, inventory, and order fulfillment processes. In particular, integer forecasts are essential to avoid rounding errors in low-volume series and to ensure operational feasibility across product portfolios with highly variable demand scales. We therefore propose a mixed-integer linear programming reconciliation (REC-MILP) that finds the reconciled point closest to the base forecast in a level/series-weighted $L_1$ distance, while restricting the solution to the integer-coherent feasible set:

\begin{equation}\label{eqn:obj}
    \min_{b_{jt} \in \mathbb{Z}_{\geq 0}, z_{it} \in \mathbb{R}_{\geq 0} } \frac{1}{NH} \sum_{i \in \mathcal{I}} \sum_{t \in \mathcal{T}} w_{i} z_{it},
\end{equation}
s.t.
\begin{equation}\label{eqn: const1}
     z_{it} \geq  \sum_{j \in \mathcal{B}} s_{ij}b_{jt} - \hat{y}_{it}
     \quad , \forall i \in \mathcal{I}; t \in \mathcal{T},
\end{equation}
\begin{equation}\label{eqn:const2}
     z_{it} \geq  \hat{y}_{it} - \sum_{j \in \mathcal{B}} s_{ij}b_{jt}
     \quad , \forall i \in \mathcal{I}; t \in \mathcal{T}.
\end{equation}

Let $\mathcal{I}$ denote the set of all series across all hierarchy levels ($|\mathcal{I}|=N$), $\mathcal{B}$ the set of bottom-level series ($|\mathcal{B}|=B$), and $\mathcal{T} = \{T+1, \dots,T+H\}$ the forecast horizons. Decision variables are bottom-level reconciled forecasts $b_{jt} \in \mathbb{Z}_{\geq 0}$. Absolute deviations from the base forecasts are captured by auxiliary variables $z_{it} \in \mathbb{R}_{\geq 0}$. Objective (\ref{eqn:obj}) minimizes the sample-average of weighted absolute errors, so REC-MILP can be read as empirical risk minimization over the class of integer-coherent forecasts. Constraints (\ref{eqn: const1})--(\ref{eqn:const2}) linearize the absolute error term and reconciled values at any level follow from bottom-level decisions via $\tilde{y}_t = (\mathbf{S}b_t)_i$. The weights $w_{i}$ control how strongly deviations from the base forecast are penalized. 
In our baseline REC-MILP, we set $w_{i}=\gamma_i$ to reflect series-specific reliability (e.g., $\gamma_i = 1/WMAPE_i$ or other accuracy-based proxy). A higher $\gamma_i$ penalizes deviations from reliable base forecasts more strongly, preserving their accuracy. Conversely, a lower $\gamma_i$ allows greater adjustments in less trustworthy forecasts, promoting corrections where base models underperform. In the level-weighted variant (REC-LW-MILP), we set $w_{i}= \gamma_i  \alpha_{l(i)}$, where $l(i)$ maps each series to its level $l \in \mathcal{L}$, and $\alpha_l \geq 0, \sum_{l \in \mathcal{L}} \alpha_l = 1$, encode managerial priorities across the hierarchy. For instance, demand at higher aggregation levels (e.g., brand or market) carries greater managerial importance, whereas lower levels (e.g., product type) are critical for operational feasibility though typically noisier. This level-weighted design provides flexibility by allowing decision-makers to emphasize different hierarchy levels depending on the planning task.

\section{Case study}
\subsection{Data}\label{sec: data}
The dataset was provided by a German premium OEM and comprises approximately 190,000 monthly observations of actual orders for a single brand. The observation period covers 12 years in total. The training set spans Years 1–11, the validation set (used for hyperparameter optimization) covers Years 9–11, and the test set corresponds to Year 12. The dataset is detailed at the market-product type level, encompassing 4,386 time series across 23 markets and 501 product types. These time series can be aggregated across a four-level product hierarchy, which is defined by Brand, Product Cluster, Product Line, and Product Type (Table \ref{Tab: Data Level}). 

\begin{table}[ht!]
\centering
\resizebox{0.9\textwidth}{!}{
\setlength\extrarowheight{-4pt}

\begin{tabular}{l l l l}
\toprule 
    ID & Level & Description & \# Time series \\ \hline 
    0 & Brand & Total brand orders & 1\\  
   1 & Market & Total orders, aggregated for each market & 23\\
2 & Market--Product Cluster & Total orders, aggregated for each market and product cluster& 115\\
3 & Market--Product Line & Total orders, aggregated for each market and product line& 412\\
4 & Market--Product Type & Total orders, aggregated for each market and product type& 4,386\\
   
\bottomrule

\end{tabular}
}
\caption{Number of time series per hierarchy level and corresponding level descriptions.}\label{Tab: Data Level}
\begin{minipage}{\textwidth}
        \footnotesize
        \textit{Notes.} High levels of aggregation correspond to low identification numbers (IDs) (e.g., levels 0 and 1), while low levels of aggregation correspond to higher IDs (e.g., levels 2 and 3), with level 4 being the most granular level. 
    \end{minipage}
\end{table}

Within the dataset, each product type is characterized by several attributes (e.g., body type, fuel type, transmission, performance, number of doors and cylinders, and life cycle age). These attributes will henceforth be referred to as master data. In addition to the master data, the set of explanatory variables is extensive, incorporating publicly available macroeconomic data extracted from Eurostat and stock market data retrieved from Yahoo Finance (see Table EC.4 of e-companion EC.6.1), as well as company-internal information such as configurator website visits, inventory levels, sales targets, and pricing-related variables. 

The forecasting horizon was set to six months to align with the company’s short- to medium-term planning cycles. This supports operational and tactical decisions such as adjusting production, managing inventory, and monitoring alignment with sales targets, while providing sufficient lead time for supplier and logistics coordination. We employed a six-month rolling horizon forecasting approach,
retraining the models at each period, resulting in seven test windows.

\subsection{Exploratory data analysis}
\textbf{Spatiotemporal data characteristics across hierarchical levels.} Table \ref{Tab: summary stats ts} shows that the share of stationary series increases with granularity, reaching nearly 46\% at the Market–Product Type level. Stationarity was assessed using the Augmented Dickey-Fuller test, applied individually to each time series. The Brand level aggregate displays strong non-stationarity, capturing long-term trends and seasonality. In contrast, at the lowest aggregation level, nearly half of the time series show stationarity. This shift in stationarity suggests that forecasting models must adapt their assumptions regarding trends, differencing requirements, and error structures by hierarchical level (RQ1). At aggregated levels, models should emphasize trend and seasonal components, while at lower levels, capturing local variability is more important.

Autocorrelation was tested using the Breusch-Godfrey test for both first- and third-order serial correlation. The analysis reveals strong temporal dependencies: the single Brand-level series in our sample exhibited autocorrelation up to lag 3, while at the Market level nearly 87\% of series did so. At the Market–Product Type level, about 58\% showed autocorrelation at this order. Partial autocorrelations at lags 1, 2, and lag 13 suggest both short-term persistence and annual seasonal effects (see Figure EC.4 of e-companion EC.6.3). Forecasting approaches need to incorporate these short-term and seasonal dependencies explicitly to improve accuracy (RQ1, RQ2).

Spatial correlations, examined via the cross-correlation function, reveal that demand fluctuations in different markets are synchronized at lag 0, indicating simultaneous influence by shared external drivers such as economic shifts or marketing campaigns (see Figure EC.4 of e-companion EC.6.3). This spatial interaction must be modeled to capture cross-market dependencies and improve hierarchical forecasting (RQ1, RQ2).

\begin{table}[H]
\begin{adjustbox}{width=0.9\textwidth,center=\textwidth}
\setlength\extrarowheight{-4pt}

\begin{tabular}{llllllll}
\toprule
Level                    & Stationarity 
& \multicolumn{2}{l}{Autocorrelation 
} & \multicolumn{4}{l}{Demand classification 
} \\ \cline{3-8} 
                         &            & First-order  & $\leq$ Third-order & Smooth  & Intermittent  & Erratic  & Lumpy \\
\hline
0: Brand                 & 0.00       & 100.00     & 100.00                 & 100.00  & 0.00          & 0.00     & 0.00  \\
1: Market                & 39.13      & 86.96      & 100.00                 & 73.91   & 0.00          & 26.09    & 0.00  \\
2: Market--Product Cluster & 37.39      & 73.04      & 79.13                  & 42.60   & 0.00          & 57.40    & 0.00  \\
3: Market--Product Line         & 36.89      & 56.79      & 62.86                  & 28.15   & 0.00          & 71.85    & 0.00  \\
4: Market--Product Type     & 45.96      & 56.49      & 58.23                  & 20.70   & 0.00          & 79.30    & 0.00  \\
\bottomrule
\end{tabular}
\end{adjustbox}

\caption{Percentage of time series per level by stationarity, autocorrelation, and demand classification.}\label{Tab: summary stats ts}
\vspace{0.5em}
\end{table}
    
\textbf{Demand classification and variability.} Applying the \cite{Syntetos2005} framework shows that demand becomes smoother at higher aggregation levels, while erratic behavior increases considerably at more granular levels (Table \ref{Tab: summary stats ts}). At the lowest level, nearly 80\% of series exhibit erratic demand patterns. Intermittent and lumpy demand patterns are not present in the data. Erratic demand at granular levels reflects diverse customer choices and localized market effects, increasing forecasting complexity and uncertainty. This aligns with existing literature reporting erratic demand patterns in manufacturing products \citep{Geng2009, Onyeocha2015}. Hence, models must accommodate this variability, possibly through robust error structures or probabilistic approaches, to maintain forecast reliability (RQ1, RQ2).

\begin{figure}[H]
     \centering
     \begin{subfigure}[b]{0.49\textwidth}
         \centering
         \includegraphics[width=\textwidth]{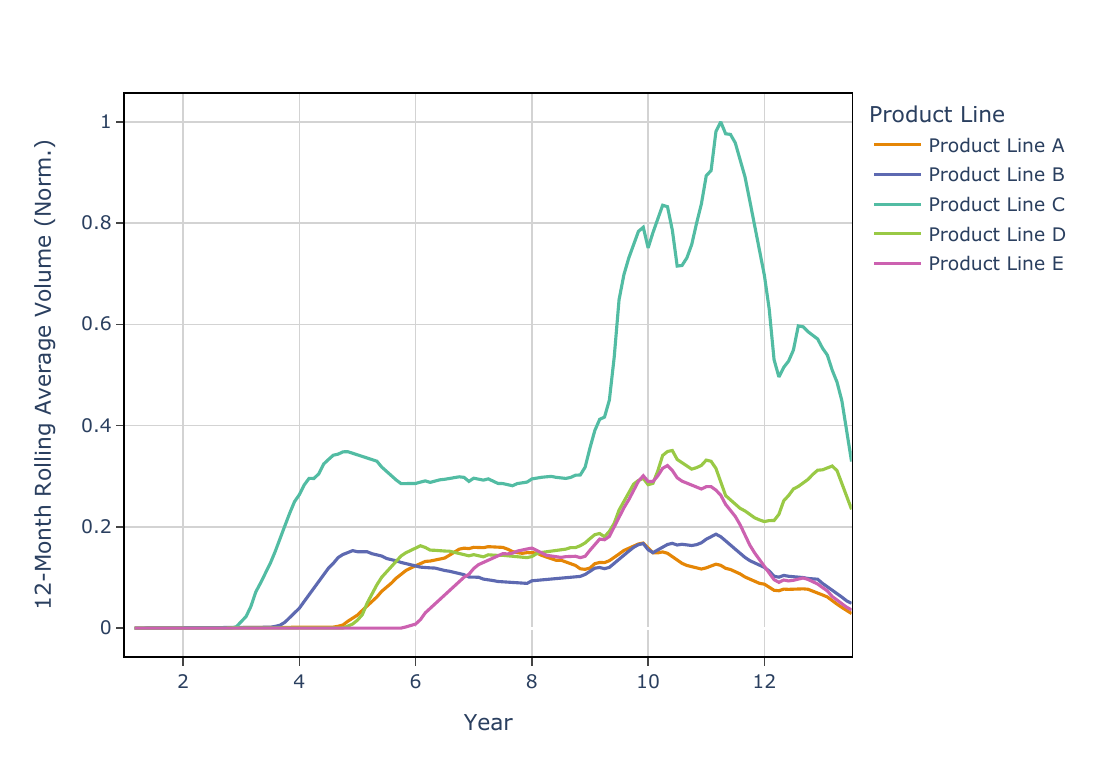}
         \caption{Life cycle patterns}\label{fig: life cycle}
     \end{subfigure}
     \begin{subfigure}[b]{0.49\textwidth}
         \centering
         \includegraphics[width=\textwidth]{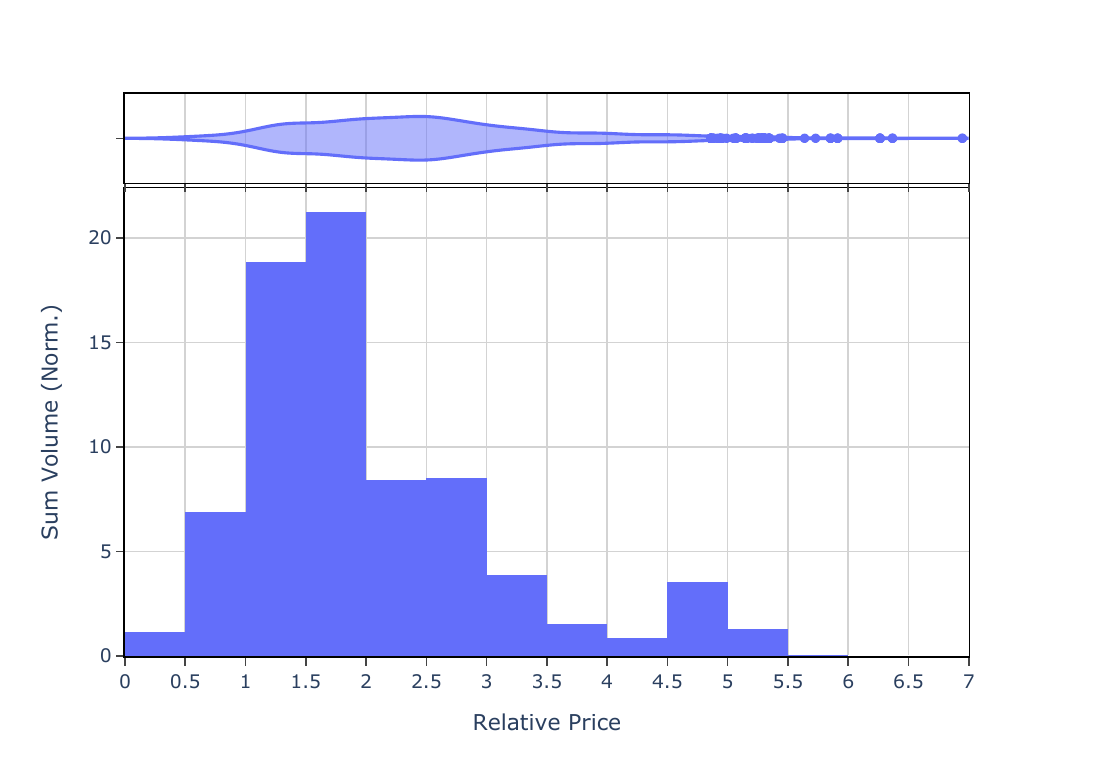}
         \caption{Pricing dynamics}
         \label{fig: pricing}
     \end{subfigure} 
     \begin{subfigure}[b]{0.49\textwidth}
         \centering
         \includegraphics[width=\textwidth]{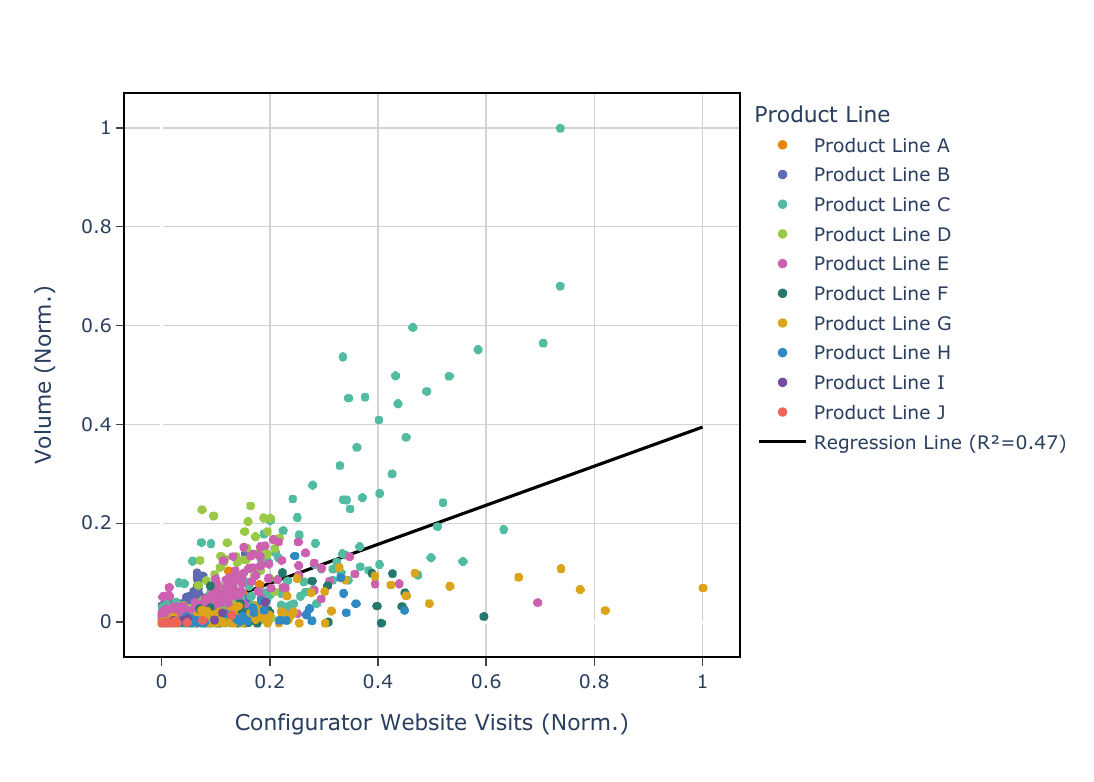}
         \caption{Configurator website visits}
         \label{fig: webkpi}
     \end{subfigure}
        \caption{Summary plots for exploratory data analysis (Market-Product Line level).}
        \label{fig: eda plots}
\end{figure}

\textbf{Product life cycles.} Figure \ref{fig: life cycle} presents life cycle volumes at the Market-Product Line level. Life cycles typically last 5–7 years, though with notable deviations from classical PLC models. Instead of a smooth introduction-growth-maturity-decline trajectory, the data reveal multiple demand peaks during maturity phases, linked to product innovations and technical updates. The complex, multi-peak life cycle patterns indicate that demand is strongly driven by product innovation timing and market acceptance dynamics rather than simple linear progression. Thus, accurate forecasting requires models that integrate product life cycle stages as core predictors (RQ2).  
\newpage
\textbf{User-generated online information.} In the automotive industry, which is dominated by wholesale-centered, indirect sales models, OEMs have limited visibility into actual customer purchasing behavior. User-generated data, such as configurator website visits, therefore provides valuable customer- and product-specific signals that would otherwise be unavailable. Regression analysis shows that configurator visits explain about 47\% of demand variance at the Market--Product Line level (Figure \ref{fig: webkpi}) but only 30\% at the Market--Product Type level. Hence, car configurator visits provide valuable signals that can enhance demand forecasts (RQ3), improving early detection of demand shifts and supporting responsive marketing decisions. 

\textbf{Price influence on demand.} Analysis of relative pricing (Market--Product Line average price divided by Market--Product Cluster average price) against cumulative volume (Figure \ref{fig: pricing}) reveals distinct demand peaks between price ratios of 1.5–2.0, indicating a perceived optimal value range. Demand declines below 1.5 and above 2.0, reflecting customer behavior with possible quality concerns or due to price sensitivity. A secondary demand peak appears between 4.5–5.0, suggesting premium segment behavior, where customers are willing to pay a premium for perceived quality, exclusivity, or brand prestige, reflecting behaviors associated with Veblen goods. Thus, price sensitivity and segment-specific demand elasticity are influential explanatory variables for automobile demand forecasting (RQ2).

\subsection{Benchmark models}\label{sec: benchmark models}
We benchmark DRFAM-PP against both statistical and ML-based methods. All models are trained separately for each quantile. Due to data sparsity at higher aggregation levels, the following methods are applied only at the Market--Product Type level: DRFAM-PP (with pooled LightGBM base models), DeepAR \citep{Salinas2020}, and NBEATSx \citep{Olivares2023}. Importantly, we run two DRFAM-PP variants: i) hierarchical pooling, using the exogenously defined hierarchy with pools at the Market, Market–Product Cluster, and Product Line levels (DRFAM-PP (HIER)); and (ii) optimal pooling via POOL-SEL-BP, which endogenously selects pools (DRFAM-PP (OPT)). For the latter, the selected pools are body type, market, and market cluster (see e-companion EC.3 for selection details). All other forecasting  strategies are implemented across the full hierarchy. Specifically, DIR, REC, HYB, and their ensemble ENS are implemented using global LightGBM models. In addition, we include six statistical baselines: naive, seasonal naive (sNaive), moving average (MA), exponential smoothing (ETS), ARIMA, and Prophet \citep{Taylor2018}. 

\subsection{Feature engineering, data processing, and model tuning}
Feature engineering extends baseline covariates with two domain-neutral interaction terms: (i) the age–volume moment (AVM), linking demand to product maturity, and (ii) a web-traffic–demand interaction (WDI), capturing the joint intensity of digital attention and demand. To prevent leakage, both use a one-sided $k$-period smoother $S_k(\cdot)$ computed with information up to $t-1$; in our implementation, we set $k=3$ and use a trailing moving average. For series $i$ at time $t$ with introduction date $s_i$, let age $a_{it} = t-s_i$. The AVM is $AVM_{it} = a_{it} S_k(y_{i, \leq t-1})$. AVM is additive across the hierarchy, aggregating by simple summation to level $l$: $AVM_{lt} = \sum_{i\in\mathcal{I}: l(i)=l} AVM_{it}$. Let $v_{it}$ denote online visits for series $i$ at time $t$. The WDI is $WDI_{it} = S_k(v_{i, \leq t-1}) S_k({y_{i,\leq t-1}})$ and aggregates analogously: $WDI_{lt} = \sum_{i \in \mathcal{I}: l(i)=l} WDI_{it}$.

We preselect features using the Maximum Relevance Minimum Redundancy algorithm \citep{Ding2003}, trained on mutual information, to identify $m=40$ explanatory variables for each forecast horizon. Selecting 40 variables strikes a balance between capturing sufficient explanatory power and avoiding overfitting or instability in interpretation. The union of all selected features is reported in the e-companion (EC.6.2, Table EC.5). This preselection is crucial for minimizing the risk of potentially misleading Shapley Additive Explanations (SHAP) values, as high correlations among features can lead to imprecise interpretations and even opposite signs \citep{Aas2021}, contrary to the original assumption of feature independence made by \cite{Lundberg2017}. 

To address extrapolation limitations of tree-based models, we apply first-order differencing to all time series, 
and quantile-transform the target variable to approximate normality. Temporal features are encoded with one-hot variables for year and quarter, and sinusoidal transformations for month and month-of-quarter. Categorical predictors are one-hot encoded when categories are few; otherwise, high-cardinality features are handled using LightGBM’s native categorical encoding \citep{Ke2017} or frequency encoding for deep learning models. Numerical features are standardized, and two principal components are extracted to reduce dimensionality. Missing values are encoded with -1 for tree-based models, while deep learning models use forward- and backfilling strategies.

Hyperparameters are optimized with \textit{Optuna} 4.1 using four-fold cross-validation on a 2-year validation set, with each fold covering six months. The tuning process is guided by root mean squared error. Implementation details, including hardware specifications and library versions, are reported in the e-companion (EC.1); hyperparameter settings in EC.6.4 (Table EC.6).

\subsection{Forecast error assessment}\label{sec: forecast error}
To assess forecast accuracy, we adopt a two-step process similar to that used in the M5 Accuracy competition: errors are first averaged over the forecast horizon for each time series, then across all series. This process is applied at each hierarchical level. We omit revenue-based weighting, as it can distort results; volume-based precision is crucial for the OEM in tasks such as production planning.

\subsubsection{Point forecast}
Root mean squared scaled error (RMSSE) is used to compare forecast accuracy across models within each hierarchical level:
\begin{equation}
    RMSSE = \sqrt{\frac{\frac{1}{H} \sum_{t=T+1}^{T+H} (y_t - \hat{y}_t)^2}{\frac{1}{T-1}
    \sum_{t=2}^T (y_t - y_{t-1})^2}}.
\end{equation}

RMSSE is scale-independent, allowing fair comparison between series with different magnitudes
or volatility; it is symmetric; and it can be computed safely, as it does not depend on values that may be equal to or close to zero \citep{Makridakis2022}. To complement RMSSE, weighted mean absolute percentage error (WMAPE) is used for cross-level comparison: 
\begin{equation}
    WMAPE = \frac{ \sum_{t=T+1}^{T+H}|y_t - \hat{y}_t|}{\sum_{t=T+1}^{T+H}|y_t|}.
\end{equation}

WMAPE is intuitive, expressing error as a percentage of actual values and useful for comparing aggregate accuracy across hierarchical levels. 

To detect systematic directional errors, we additionally report the (scaled) Forecast Bias (FB):
\begin{equation}
    FB = \frac{ \sum_{t=T+1}^{T+H} (y_t - \hat{y}_t) }{\sum_{t=T+1}^{T+H}y_t}.
\end{equation}
FB helps detect consistent over- or under-forecasting: a positive bias indicates underestimation,
which can lead to underproduction or understocking, and vice versa. While FB is useful for
identifying directional bias, it does not reflect the magnitude of errors and is therefore used as a
sanity check rather than a primary performance metric.

\subsubsection{Probabilistic forecast}
We employ the scaled pinball loss (SPL), which is computed for each time series and quantile as follows:
\begin{equation}
    SPL(q) = \frac{\frac{1}{H}\sum_{t=T+1}^{T+H}[(y_t-\hat{y}_t^q)q \mathds{1}\{\hat{y}_t^q\leq y_t\} + (\hat{y}_t^q-y_t)(1-q) \mathds{1}\{\hat{y}_t^q> y_t\}]}{\frac{1}{T-1}\sum_{t=2}^T|y_t - y_{t-1}|},
\end{equation}
where \(\mathds{1}\{\cdot\}\) is the indicator function that equals 1 if \(y_t\) falls within the specified interval, and 0 otherwise. Similar to the point forecast, the denominator is calculated only for periods with non-zero demand. The pinball loss function penalizes overestimation more heavily for low-probability quantiles, and underestimation more for high-probability quantiles. SPL is normalized similarly to the RMSSE, enabling consistent comparisons across time series with different scales. It is numerically stable, as it avoids divisions by values approaching zero. Moreover, to fairly estimate the distribution of forecast uncertainty, no weights are applied to the evaluated quantiles, ensuring an unbiased representation of uncertainty across the forecast distribution \citep{Makridakis2022}. The mean SPL (MSPL) is obtained by averaging over all quantiles:
\begin{equation}
   MSPL = \frac{1}{|\mathcal{Q}|} \sum_{q \in \mathcal{Q}} SPL(q).
\end{equation}

\section{Results}
\subsection{Forecast method effectiveness}
\begin{figure}[H]
     \centering
     \begin{subfigure}[b]{0.49\textwidth}
         \centering
         \includegraphics[width=\textwidth]{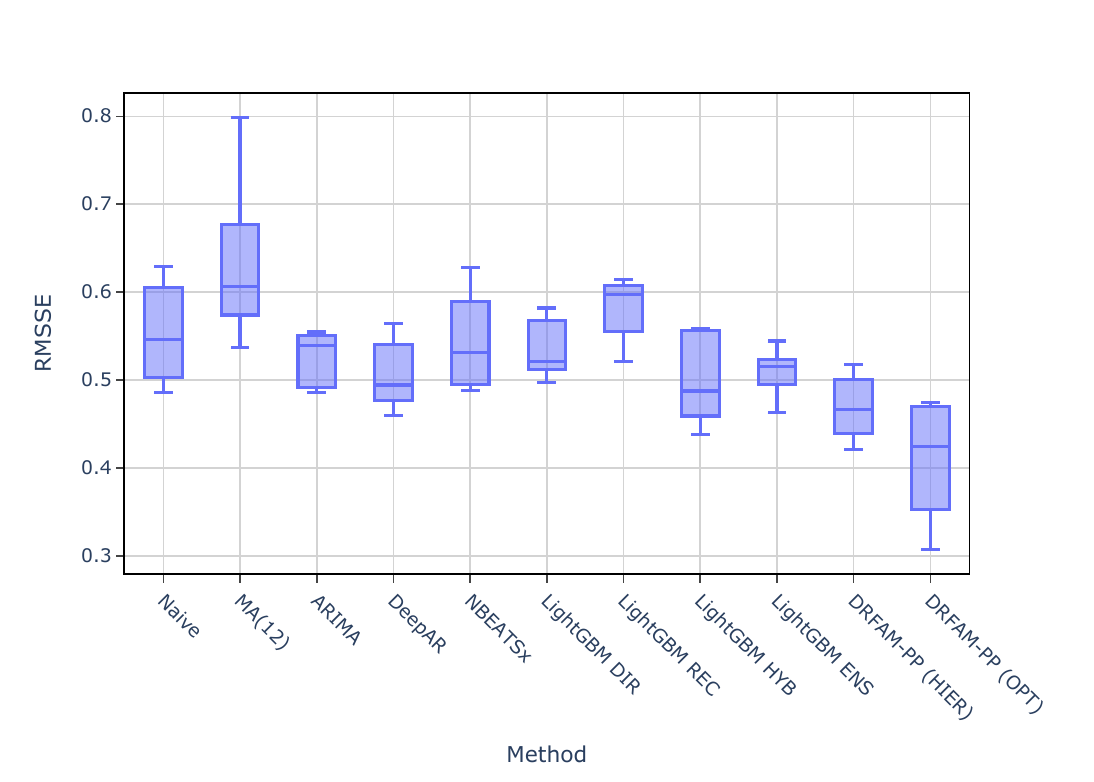}
         \caption{RMSSE}
         \label{fig: rmsse lwest level}
     \end{subfigure}
     \begin{subfigure}[b]{0.49\textwidth}
         \centering
         \includegraphics[width=\textwidth]{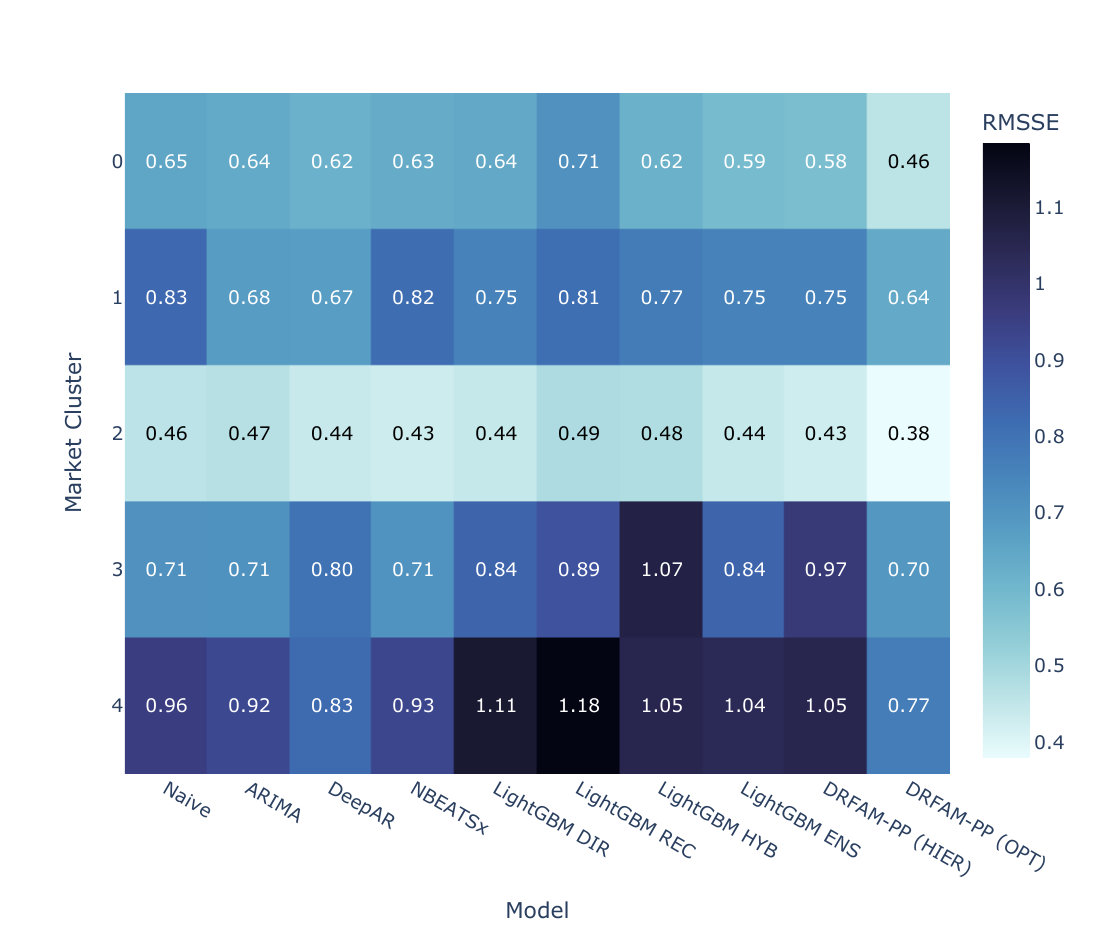}
         \caption{RMSSE by market cluster}
         \label{fig: rmsse by market}
     \end{subfigure} 

     \begin{subfigure}[b]{0.49\textwidth}
         \centering
         \includegraphics[width=\textwidth]{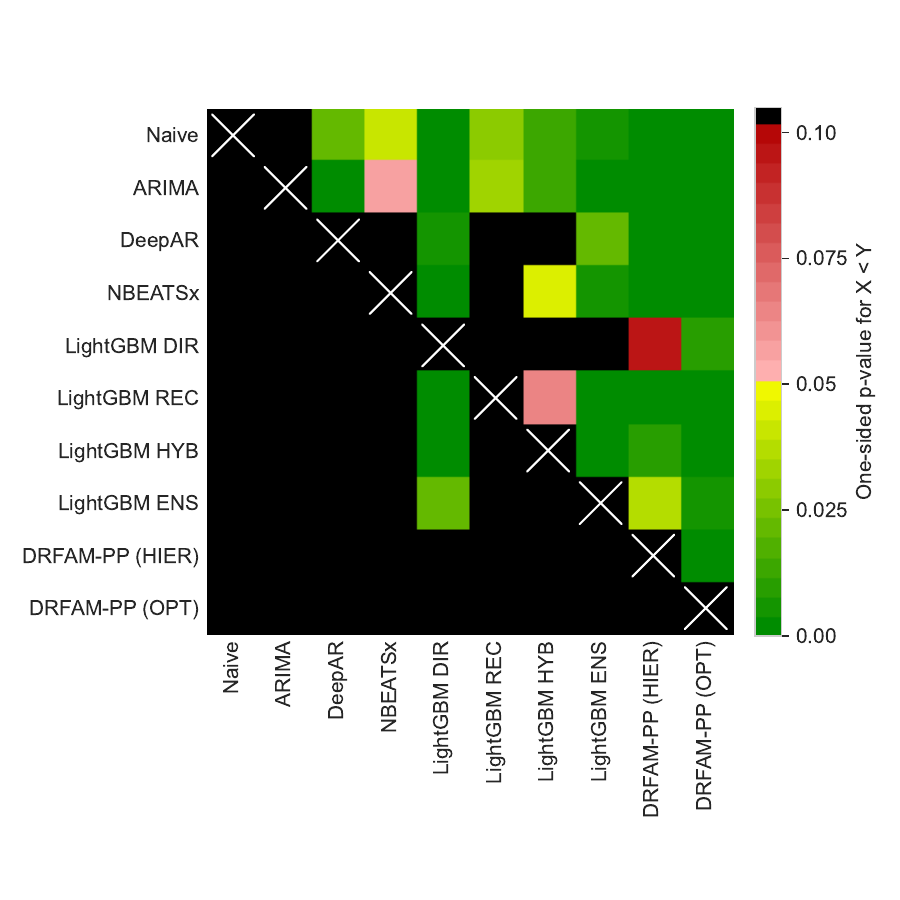}
         \caption{Robustness point forecast}
         \label{fig: forcast perf point}
     \end{subfigure}
     \begin{subfigure}[b]{0.49\textwidth}
         \centering
         \includegraphics[width=\textwidth]{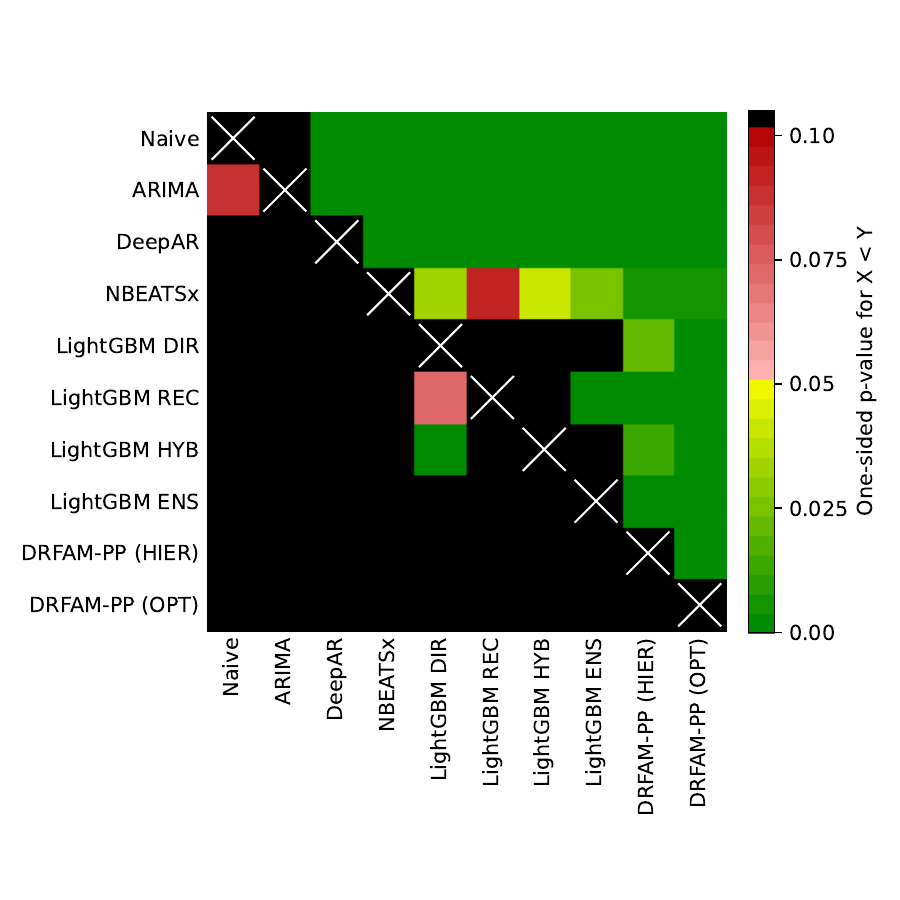}
         \caption{Robustness quantile forecast}
         \label{fig: forcast perf prob}
     \end{subfigure}
     
        \caption{Summary plots for Market-Product Type level accuracy.}
        \label{fig:accuracy plots}

\vspace{0.5em}
    \begin{minipage}{\textwidth}
        \footnotesize
        \textit{Notes.} Panels (a), (b): report median scores calculated over the seven test periods. Panels (c), (d): Giacomini–White test; HAC (Newey–West) with lag $L=H-1$; losses are MSE (c) and CRPS via trapezoidal approximation (d).
    \end{minipage}
\end{figure}

Table \ref{tab:results_overview} compares point (RMSSE) and probabilistic (MSPL) forecast accuracy across hierarchical levels from Brand to Market--Product Type, including only the best-performing statistical benchmark (full statistical benchmark results are in \ref{app:benchmarks}; extended accuracy metrics, including WMAPE and FB, are reported in the e-companion EC.5.6). The Market--Product Type level accuracy evaluation is further supported by Figure \ref{fig:accuracy plots}. 

\begin{table}[h]
\centering
\begin{adjustbox}{width=\textwidth,center=\textwidth}
\setlength\extrarowheight{-4pt}

\begin{tabular}{lllllllllll}
\toprule
Level & Metric & Statistical & DeepAR & NBEATSx & DIR & REC & HYB & ENS & \shortstack{DRFAM-PP \\ (HIER)} & \shortstack{DRFAM-PP \\ (OPT)} \\ \hline
0: Brand & RMSSE & \underline{\textbf{0.640}} & \dag & \dag & 1.145 & 0.750 & 1.267 & 1.020 & \dag & \dag\\
& MSPL & 0.334 & \dag & \dag & 0.320 & \underline{\textbf{0.233}} & 0.292 & \textbf{0.240} & \dag & \dag \\
1: Market & RMSSE & 0.648 & \dag & \dag & 0.606 & 0.638 & 0.730 & \underline{\textbf{0.579}} & \dag& \dag \\
& MSPL & 0.362 & \dag & \dag & 0.249 & \underline{\textbf{0.202}} & 0.217 & \textbf{0.206} & \dag & \dag\\
2: Market--Product Cluster & RMSSE & 0.608 & \dag & \dag & 0.621 & 0.618 & 0.804 & \underline{\textbf{0.546}} & \dag & \dag\\
& MSPL & 0.390 & \dag & \dag & 0.229 & \textbf{0.206} & 0.257 & \underline{\textbf{0.200}} & \dag & \dag\\
3: Market--Product Line & RMSSE & 0.579 & \dag & \dag & 0.521 & 0.543 & 0.675 & \underline{\textbf{0.436}} & \dag & \dag\\
& MSPL & 0.362 & \dag & \dag & 0.185 & 0.155 & 0.206 & \underline{\textbf{0.145}} & \dag & \dag\\ 
4: Market--Product Type & RMSSE & 0.539 & 0.494 & 0.531 & 0.581 & 0.576 & 0.478 & 0.516 & 0.469 & \underline{\textbf{0.418}} \\
& MSPL & 0.310 & 0.182 & 0.654 & 0.227 & 0.171 & 0.154 & 0.181 & \textbf{0.134}  & \underline{\textbf{0.128}} \\

\bottomrule
\end{tabular} 
\end{adjustbox}

\caption{Point and probabilistic forecast accuracy across hierarchical levels.}
\label{tab:results_overview}

\vspace{0.5em}
    \begin{minipage}{\textwidth}
        \footnotesize
        \textit{Notes.} \dag: Insufficient data available for the application of the specified model. Underlined values indicate the best accuracy, while bold numbers represent values that are at most 5\% worse than the best.
    \end{minipage}
\end{table}

The main results related to RQ1 are summarized below (minor results are in e-companion EC.5). Brand level: simple statistical benchmarks yield the lowest RMSSE (0.640), consistent with stable aggregate patterns. LightGBM REC and ENS variants deliver the best probabilistic scores, reflecting their superior ability to quantify forecast uncertainty. Intermediate levels (1–3): ENS attains the best point and probabilistic accuracy (improvements: RMSSE--15.24\%, SPL--50.59\% vs. best statistical baseline). Market–Product-Type level: DRFAM-PP (OPT) outperforms all benchmark methods in both point (Figure \ref{fig: rmsse lwest level}) and probabilistic (see e-companion EC.6.6) accuracy, followed closely by LightGBM HYB and ENS variants. Robustness checks using Giacomini–White tests confirm these rankings (Figures \ref{fig:accuracy plots}c,d). Moreover, hierarchical pooling is empirically non-optimal: the tests favor optimal, data-driven pooling over fixed hierarchical pooling (DRFAM-PP (HIER)) in pairwise comparisons. Deep learning models (DeepAR, NBEATSx) perform competitively at level 4 but exhibit long-horizon overprediction bias (see e-companion EC.6.5), whereas both DRFAM-PP variants maintain stable bias across all forecast horizons and test periods.



\textbf{Finding 1: Spatial heterogeneity of forecast performance.} Accuracy gains from DRFAM-PP (OPT) are not spatially uniform but depend on market maturity, volume, and volatility (Figure \ref{fig: rmsse by market}). The largest improvements occur in high-volume markets (clusters 0 and 1), where the economic stakes are highest, while in saturated markets (cluster 3) improvements over naive benchmarks were marginal. In growth-phase markets, forecast errors were higher across all models due to limited history and evolving dynamics; here DRFAM-PP (OPT) improved RMSSE by 3.65\%, compared to more pronounced gains of 10.29\% in mature markets. At the individual market level, DRFAM-PP (OPT) delivered top accuracy in 65\% of cases, primarily in stable, high-sales environments. By contrast, global ensemble methods retained an edge in low-volume or rapidly evolving markets.

These patterns generalize beyond automotive to any multi-region, multi-segment demand system: forecasting performance varies with spatial structure and market phase, not just hierarchical level. A context-dependent deployment is therefore preferable to a uniform rollout: allocate resource-intensive models like DRFAM-PP (OPT) to mature, high-impact regions, and use faster-to-implement, adaptive global models where data are sparse or dynamics shift quickly. This suggests a hybrid strategy where model choice becomes context-dependent, aligning forecasting investment with economic return.

\textbf{Finding 2: Rounding substantially impacts forecast accuracy, and MILP-based reconciliation outperforms benchmark methods.} We compared REC-MILP and its level-weighted extension (REC-MILP-LW) against benchmark reconciliation methods (BU, MinT) and unreconciled base forecasts from ETS, ARIMA, and LightGBM REC. For BU and MinT, reconciliation was applied to continuous forecasts with post hoc rounding. Detailed methodological explanations and complete result tables are provided in \ref{app: reconciliation}.

\begin{figure}[H]
     \centering
     \begin{subfigure}[b]{0.49\textwidth}
         \centering
         \includegraphics[width=\textwidth]{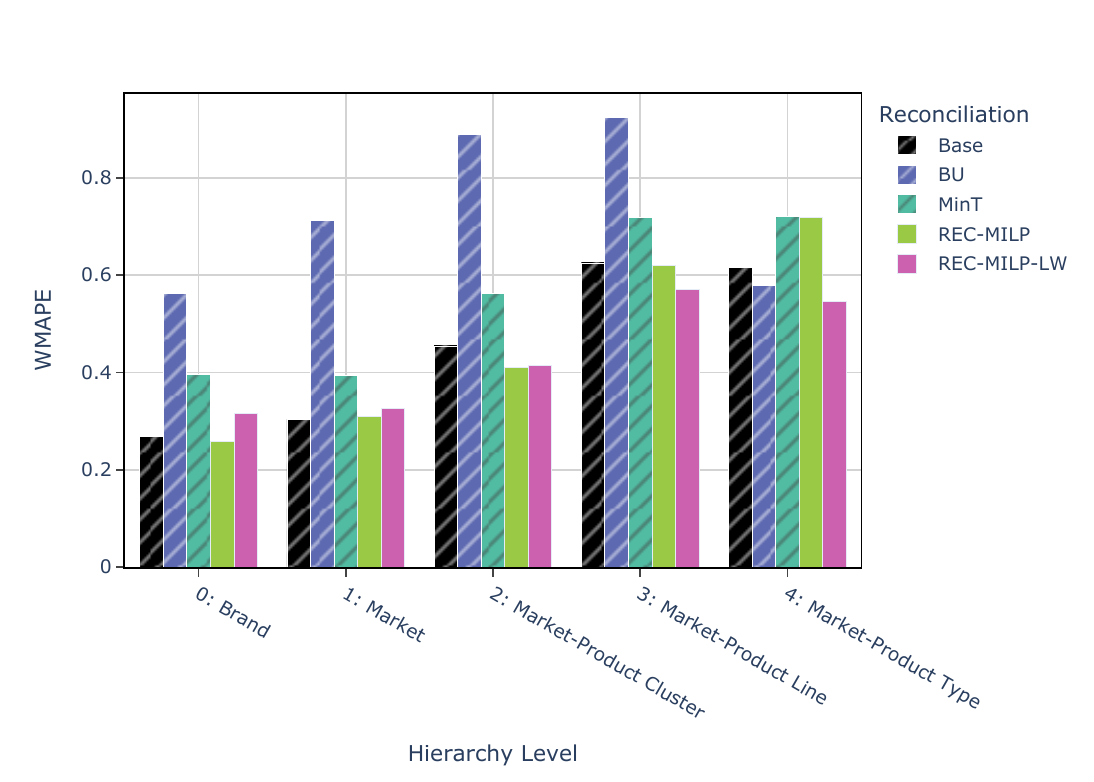}
         \caption{Cross-level accuracy}\label{fig: rec levels}
     \end{subfigure}
    \begin{subfigure}[b]{0.49\textwidth}
         \centering
         \includegraphics[width=\textwidth]{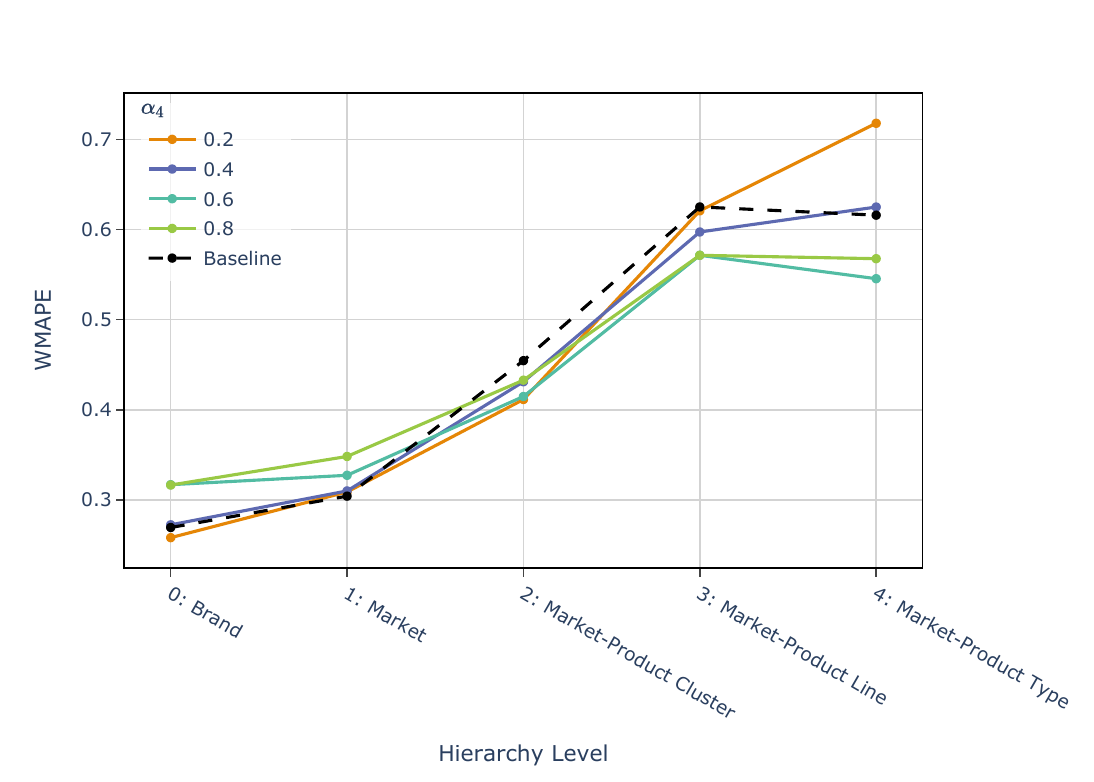}
         \caption{Level-weight sensitivity}\label{fig: rec lambda sensitivity}
     \end{subfigure}
        \caption{Forecast accuracy of reconciliation methods across hierarchical levels.}\label{fig: rec sensitivity}

\vspace{0.5em}
    \begin{minipage}{\textwidth}
        \footnotesize
        \textit{Notes.} Figures show median WMAPE scores for the ARIMA base forecast. In panel (a), shaded bars indicate reconciliation methods that are non-coherent after integer-rounding.
    \end{minipage}
\end{figure}

Results show that rounding can meaningfully distort accuracy: at the Market–Product Type level (Figure \ref{fig: rec levels}), rounding improved BU relative to unrounded base forecasts but generally increased error for BU and MinT at other levels (\ref{app: reconciliation}, Tables \ref{tab: rec rmsse}--\ref{tab: rec wmape}). By contrast, REC-MILP consistently avoided this degradation and in many cases matched or improved base forecast accuracy. The level-weighted variant (REC-MILP-LW) further demonstrated that targeting the objective to the Market–Product Type level improved bottom-level accuracy with only modest trade-offs at higher levels (Figure \ref{fig: rec lambda sensitivity}; \ref{app: reconciliation}, Table \ref{tab: rec wilcoxon}). Gains were strongest for ARIMA and LightGBM REC, where variable forecasts across the horizon are more sensitive to rounding bias. In contrast, ETS produces constant forecasts over the horizon, limiting rounding effects; differences were small and statistically insignificant.

From an applied forecasting perspective, these results underscore two implications. First, reconciliation is not only about achieving hierarchical coherence but also about operational implementability: integer bottom-level forecasts matter wherever plans translate directly into discrete production, inventory, or fulfillment decisions. Second, REC-MILP-LW offers practitioners a practical tuning mechanism: managers can assign weights to emphasize operational feasibility (e.g., production scheduling) or strategic alignment (e.g., brand/region planning). Importantly, the method delivers robustness as it avoids accuracy losses where rounding has little impact, while producing the largest gains in settings where variance and rounding sensitivity are high. This makes integer-coherent reconciliation particularly valuable for industries characterized by low volumes, high product variety, or discrete production and allocation decisions (e.g., automotive, electronics, pharmaceuticals, and retail), where precise bottom-level forecasts help avoid costly overproduction, lost sales, or inefficiencies in sequencing and replenishment processes. In addition to its accuracy benefits, REC-MILP(-LW) also demonstrates competitive computational efficiency. In our study, MinT required approximately one hour to compute the reconciliation using a full shrinkage-based covariance estimate, while REC-MILP solved the integer optimization in seconds. This efficiency extends to solving reconciliation across full hierarchies and horizons simultaneously, making the method scalable to large industrial forecasting systems.

\subsection{Predictors of automobile demand}
\begin{figure}[H]
     \centering
     \begin{subfigure}[b]{0.68\textwidth}
         \centering
         \includegraphics[width=\textwidth]{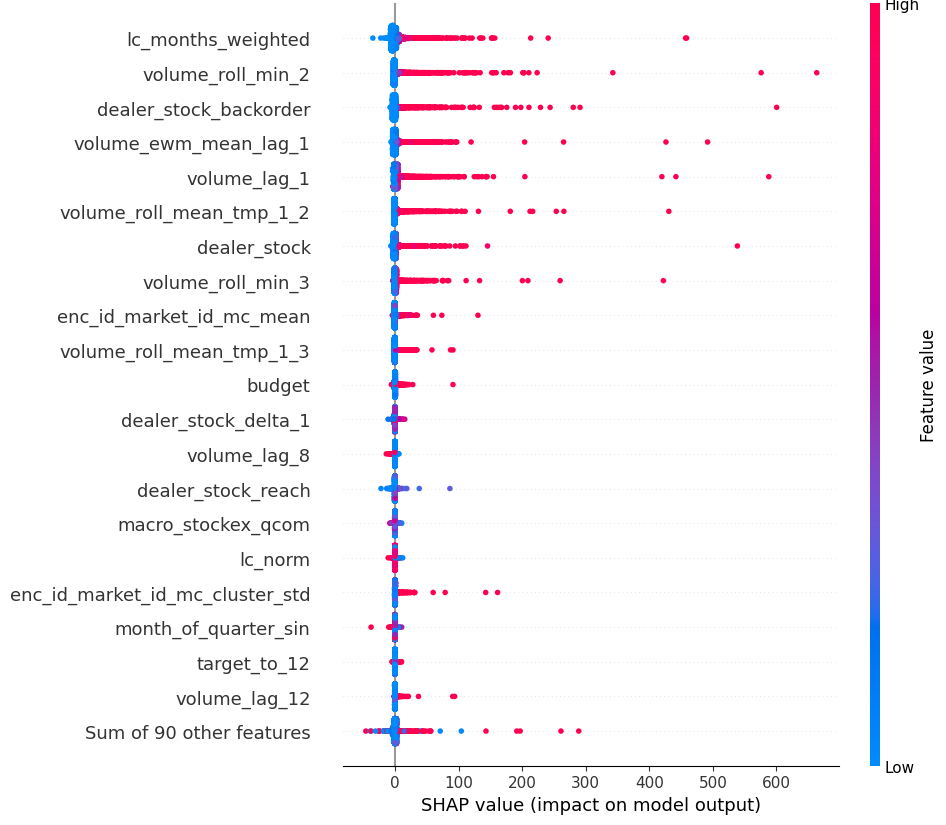}
         \caption{$h=1$}\label{fig: shap lag 1}
     \end{subfigure}
     \begin{subfigure}[b]{0.68\textwidth}
         \centering
         \includegraphics[width=\textwidth]{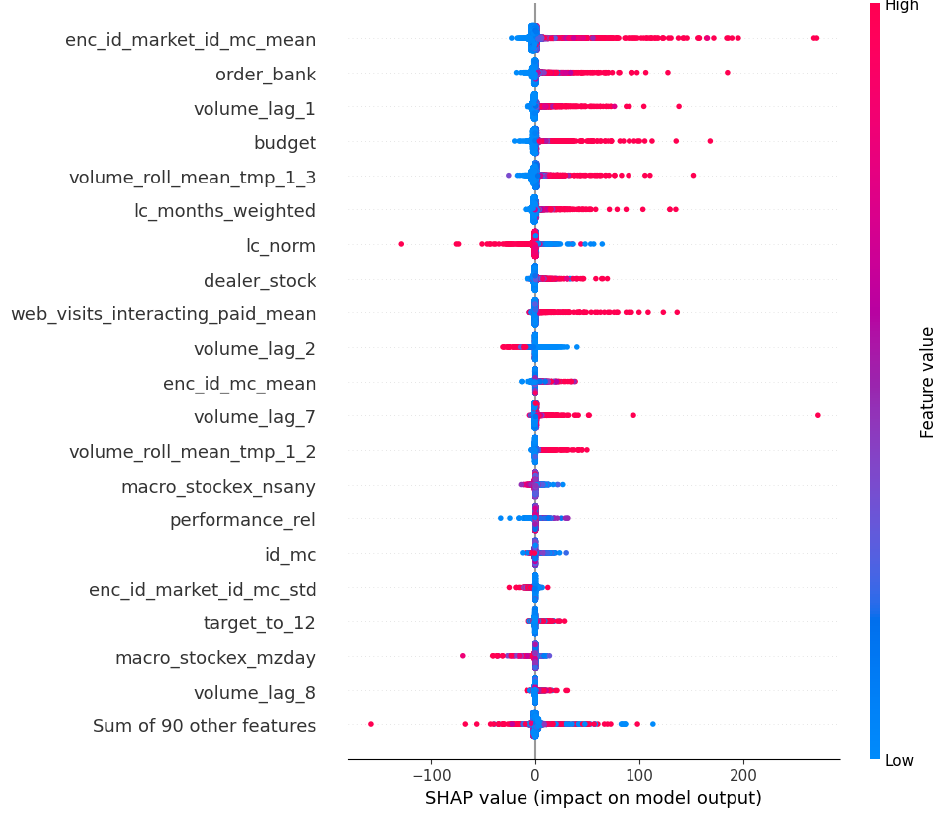}
         \caption{$h=6$}
         \label{fig: shap lag 6}
     \end{subfigure} 
        \caption{Distribution of SHAP values per feature for forecast steps 1 and 6.}
        \label{fig: shap plots}
\end{figure}

We conducted a global SHAP analysis to identify the most influential predictors of automobile demand (RQ2) at the Market--Product Type level. SHAP values were computed using the TreeSHAP algorithm. Figure \ref{fig: shap plots} presents the distribution of SHAP values per feature for the 1-month and 6-month forecast horizons based on a direct LightGBM model. Although this model did not yield the highest accuracy in earlier comparisons, it offers strong interpretability, allowing us to examine how individual input features contribute to demand predictions across different forecast horizons. Minor results corresponding to RQ2 are reported in the e-companion (EC.5.5).

\textbf{Finding 3: Life cycle dynamics, recent demand patterns, and operational signals are key predictors of short-term automotive demand.} At short-term horizons (e.g., one month ahead), the strongest predictors of automotive demand are (i) a volume-weighted life cycle variable, indicating sensitivity to model maturity, (ii) autoregressive sales features (rolling means and exponentially weighted averages), and (iii) operational indicators such as backorders and dealer stock levels. The life cycle effect highlights that demand is lowest for newly launched models and peaks in mid-life phases, while autoregressive features confirm that short-term demand is strongly path-dependent. Operational signals provide additional explanatory power: backorders proxy unmet demand, while low dealer stocks indicate constrained supply. Interestingly, the stock price of Qualcomm (ticker symbol: QCOM), a major semiconductor supplier, emerges as an informative external predictor, reflecting supply chain dependencies. By contrast, online configurator data contributes little at this horizon, consistent with its forward-looking nature and the lag between online interest and final purchase decisions (e.g., test drives, dealer negotiations, financing).

These results suggest that short-term automotive demand is primarily reactive, driven by recent momentum and immediate constraints rather than forward-looking signals. For practitioners, the implications are threefold: (i) incorporating life cycle maturity into models supports more accurate production scheduling during new model launches and end-of-life transitions, (ii) monitoring supplier-linked financial indicators can serve as early warnings of upstream supply chain disruptions, and (iii) embedding stock-related operational signals enhances dealer replenishment and reduces the risk of stockouts. These insights underscore the importance of combining internal operational data with external signals to improve short-term forecast robustness.

\textbf{Finding 4: Online engagement, internal sales expectations, and evolving product life cycle feature gain relevance at longer horizons.} At medium-term horizons (e.g., six months ahead), SHAP analysis reveals a distinct shift in predictor relevance compared to the short term. Internal planning targets (e.g., budget, committed orders), normalized life cycle features, and online engagement signals dominate, while the influence of weighted life cycle indicators declines. The rise of the normalized life cycle variable suggests that the model generalizes maturity effects across the portfolio, enabling inference even for early-stage or low-volume models. Online engagement data, though weak at short horizons, gains predictive power as early consumer interest gradually converts into orders. Competitive financial indicators, such as the stock performance of peer brands, show a negative influence on demand predictions, reflecting competitive pressures and potential market share shifts.

These results indicate that forecasting at longer horizons transitions from reactive signals to anticipatory drivers aligned with strategic planning. For OEMs, the implications are threefold: (i) generalized life cycle modeling enhances demand predictability for niche products, reducing reliance on sparse historical data; (ii) integrating online engagement provides early visibility into emerging consumer preferences, supporting proactive marketing and product mix optimization; and (iii) monitoring financial signals of rival brands acts as a proxy for competitive dynamics, enabling OEMs to anticipate market pressure and adapt inventory, pricing, and resource allocation accordingly. These insights highlight the value of blending internal expectations, external demand signals, and competitive intelligence in medium-term planning.

\subsection{Value of user-generated online information}
To quantify the impact of user-generated online information on forecast accuracy (RQ3), we estimated linear mixed-effects models. This modeling approach captures intra-market and intra-period dependencies arising from the
rolling-horizon evaluation procedure, where overlapping test sets introduce correlation between observations. The full specification of the mixed-effects models and statistical tests is provided in \ref{apx: mixed model}, and the results are summarized in Table \ref{tab:model_comparison}.

\begin{table}[h]
\centering
\begin{adjustbox}{width=0.85\textwidth,center=\textwidth}
\setlength\extrarowheight{-4pt}
\begin{tabular}{@{}lllllllll@{}}
\toprule
Model & Level ID & $\beta_0$ & $\beta_1$ & $\hat{\sigma}_u$ & $\hat{\sigma}_v$ & $\hat{\sigma}$& $R^2$
&$r$
\\
\midrule
LightGBM DIR     & 4   & 1.015 &–0.022 [-0.044, -0.000]\textsuperscript{*} &  0.672  & 1.623 & 0.321& 0.984 &0.714\textsuperscript{***}  \\
                    & 3    & 0.862  & –0.012 [-0.022, -0.002]\textsuperscript{*}& 0.489   & 0.887  & 0.093& 0.996 &0.495\textsuperscript{*} \\
& 0-2 & \multicolumn{7}{l}{\textit{No significant effect detected.}} \\

LightGBM REC  & 4 & 1.229 & –0.010 [-0.019, -0.000]\textsuperscript{*}&  0.821  & 3.368  & 0.144 & 0.999&0.768\textsuperscript{**}  \\
                    & 3  & 0.950    & –0.010 [-0.016, -0.004]\textsuperscript{***} &  0.525  & 1.010 & 0.057&0.999&0.635\textsuperscript{***}  \\
& 0-2 & \multicolumn{7}{l}{\textit{No significant effect detected.}} \\
DRFAM-PP (OPT)       & 4 & 0.833 & –0.058 [-0.089, -0.027]\textsuperscript{***} & 0.602  & 1.389 & 0.407& 0.965&0.121\textsuperscript{*}  \\
\bottomrule
\end{tabular} 
\end{adjustbox}

\caption{Statistical test results assessing the impact of user-generated online information on forecasting error (WMAPE) across different models and hierarchical levels.}
\label{tab:model_comparison}

 \vspace{0.5em}
    \begin{minipage}{\textwidth}
        \footnotesize
        \textit{Notes.} \textsuperscript{*}: $p<0.05$, \textsuperscript{**}: $p<0.01$, \textsuperscript{***}: $p<0.001$. \\
    \end{minipage}

\end{table}


\textbf{Finding 5: User-generated online information significantly improves forecast accuracy at disaggregated levels.} Across all evaluated forecasting models, the inclusion of user-generated online data significantly enhanced forecast accuracy at the most granular level (Market--Product Type). For example, in DRFAM-PP (OPT), the largest improvement was observed with a WMAPE reduction of 5.8\% ($p < 0.001$), confirmed by non-parametric testing (Wilcoxon signed-rank test; $r = 0.121$, $p = 0.020$). Similar gains occurred in both direct and recursive LightGBM models. At the Market--Product Line level, improvements were smaller but remained statistically significant. No benefits emerged at higher aggregation levels, largely due to the limited historical span of configurator data (available from Year 9), which restricts the overlap with long-term demand series and reduces the effective training sample. Random-effects estimates revealed substantial heterogeneity across markets and forecast periods: variance components for both consistently exceeded the residual variance. This underscores that forecast performance is shaped not only by idiosyncratic noise but also by structural market differences and temporal fluctuations, echoing the spatial dependencies highlighted in Finding 1.

The predictive value of online data is strongest when it directly captures product-level demand signals, where user activity (e.g., configurator interactions or comparable digital footprints) reflects concrete purchase intentions. For premium OEMs, integrating such forward-looking, customer-centric indicators into disaggregated forecasts supports production scheduling, inventory alignment, and market allocation. In industries with high product differentiation and direct-to-consumer digital interfaces (e.g., automotive, consumer electronics, and retail), this approach enhances responsiveness to market shifts. By contrast, at aggregated levels, online engagement data become diluted and lose predictive power, limiting their value for strategic or long-horizon projections. Similarly, generic web indicators (e.g., search-trend data) tend to capture attention at category (e.g., product line) rather than item (e.g., product type) level and are therefore only weakly related to realized demand, reducing their relevance for operational planning.

\section{Conclusion}
This study addressed automotive demand forecasting in a premium OEM setting by integrating hierarchical forecasting, life cycle modeling, spatiotemporal dynamics, and behavioral signals into a unified empirical framework.

(RQ1) Forecasting performance varies systematically by aggregation level: at intermediate levels, global ensemble models deliver the highest point and probabilistic accuracy, while at the most granular Market--Product Type level, pooled ensembles combining multiple forecasting strategies (DRFAM-PP) excel. This level is particularly critical for premium OEMs with high product variety, where accurate forecasts directly condition operational costs. Integer-coherent reconciliation (REC-MILP and REC-MILP-LW) further proved essential, ensuring operational feasibility by mitigating rounding biases that otherwise distort bottom-level forecasts. (RQ2) The analysis shows that forecast drivers evolve with the horizon. Short-term demand is dominated by reactive dynamics (e.g., life cycle maturity, autoregressive momentum, and operational constraints), while medium-term demand relies more on anticipatory signals such as online engagement, planning targets, and competitive financial indicators. (RQ3) User-generated online data significantly improves accuracy at disaggregated levels, highlighting its role as a forward-looking, customer-centric signal, though its predictive power diminishes at higher levels, where aggregation dilutes behavioral information. Importantly, random-effects estimates show that these gains depend on spatiotemporal differences.  

Beyond these findings, we makes several generalizable methodological contributions. We extend DRFAM-PP (M5 Accuracy winner) to a stochastic pooled ensemble and generalizes pooling-set selection as a data-driven partitioning problem (POOL-SEL-BP) that trades off accuracy and model complexity. Empirically, purely hierarchical pooling is non-optimal; optimal data-driven pooling yields statistically significant accuracy gains. Integer-coherent reconciliation (REC-MILP/REC-LW-MILP) unifies empirical-risk minimization with discrete feasibility and can be extended to a wide range of hierarchical and operational forecasting settings. The proposed life-cycle feature (AVM) is shape-agnostic and transferable, supporting applications beyond the automotive domain.

For practitioners, the findings provide actionable managerial guidance: deploy resource-intensive pooled ensembles in mature, high-volume markets; rely on statistical methods in growth-phase regions; integrate online engagement data into disaggregated planning; and apply discrete reconciliation to ensure forecasts translate into feasible operational decisions. Future research should extend discrete forecast reconciliation to probabilistic settings and decision-aware objectives, exploit longer histories of behavioral data, and test the framework in other hierarchical industries.  

\appendix

\section{Statistical benchmarks}\label{app:benchmarks}

\begin{table}[h]
\centering
\begin{adjustbox}{width=\textwidth,center=\textwidth}
\setlength\extrarowheight{-4pt}

\begin{tabular}{lllllllllll}
\toprule
Level & Metric & Naive & sNaive & MA(2) & MA(3) & MA(6) & MA(12) & ETS & ARIMA & Prophet \\ \hline

0: Brand & RMSSE & \underline{\textbf{0.640}} &  1.029 & \textbf{0.652} & \textbf{0.664}& 0.832 &0.950  & 0.766 & 0.982 & 1.199\\
& MSPL & 0.361 & 0.612 & \underline{\textbf{0.334}} & 0.369 & 0.465 & 0.529 &0.427 & 0.560 & 0.695\\

1: Market & RMSSE & \textbf{0.650} & \textbf{0.680} & 0.687 & \underline{\textbf{0.648}} & 0.710 & \textbf{0.658} & \textbf{0.670} & 0.685 & 0.743 \\
& MSPL & \textbf{0.372} & 0.400 &\textbf{0.379} & \textbf{0.369} & 0.432 & \textbf{0.380} & \underline{\textbf{0.362}} & 0.423 & 0.465 \\ 

2: Market--Product Cluster & RMSSE & \textbf{0.629} & 0.828 & \textbf{0.609} & \textbf{0.609} & 0.673 & 0.663 & \underline{\textbf{0.608}} & \textbf{0.631} & 0.785 \\
& MSPL &  \textbf{0.406} & 0.487 & \textbf{0.391} & 0.418 & 0.423 & 0.433 & \textbf{0.406} & \underline{\textbf{0.390}} & 0.485 \\

3: Market--Product Line & RMSSE &  0.644 & 0.857 & 0.621 & \underline{\textbf{0.579}} & 0.632 & 0.657 & 0.616 & \underline{\textbf{0.579}} & 0.773 \\
& MSPL & \textbf{0.374} & 0.490 & \textbf{0.377} & \textbf{0.373} & 0.400 &  0.407 & \underline{\textbf{0.362}} & 0.535 & 0.455 \\

4: Market--Product Type & RMSSE & \textbf{0.547} & 0.862 & \textbf{0.563} & \textbf{0.559} & 0.614 & 0.606 & 0.577 & \underline{\textbf{0.539}} & 0.949 \\ 
& MSPL & \textbf{0.321} & 0.460 & 0.336 & 0.341 & 0.477 & 0.388 & 0.346 & \underline{\textbf{0.310}} & 0.549 \\

\bottomrule
\end{tabular} 
\end{adjustbox}

\caption{Point and probabilistic forecast accuracy across hierarchical levels of statistical benchmarks.}\label{tab:results_overview_statistical}

\vspace{0.5em}
    \begin{minipage}{\textwidth}
        \footnotesize
        \textit{Notes.} Underlined values indicate the best accuracy, while bold numbers represent values that are at most 5\% worse than the best.
    \end{minipage}
\end{table}

\section{Evaluation design and results for reconciliation methods}\label{app: reconciliation}
We compared REC-MILP and its level-weighted variant (REC-MILP-LW) against benchmark reconciliation methods (BU and MinT) as well as unreconciled base forecasts from ETS, ARIMA, and LightGBM REC models. The analysis was based on the first six-month test period (Year 12, Months 1–6), with forecasts generated for all hierarchy levels. For BU and MinT, reconciliation was applied to continuous base forecasts, followed by post hoc integer rounding. In contrast, REC-MILP and REC-MILP-LW directly produced integer-coherent forecasts through optimization, eliminating the need for a separate rounding step. For REC-MILP-LW, level weights ($\alpha_l$) were tuned to prioritize bottom-level accuracy. Specifically, the bottom-level weight $\alpha_4$ was optimized by grid search over $[0.0, 1.0]$ in increments of 0.1, selecting the value that minimized bottom-level forecast error (RMSSE) on the validation set (Years 9--11) using four-fold cross-validation, with each fold covering six months. The resulting optimal weights were: ARIMA – 0.6, ETS – 0.8, and LightGBM REC – 0.4. All other levels were weighted equally according to $\alpha_l = (1-\alpha_4)/4, \forall l=0,\dots,3$.

Evaluation metrics included: (i) WMAPE and RMSSE across all hierarchy levels (Tables \ref{tab: rec wmape} and \ref{tab: rec rmsse}); (ii) Hodges–Lehmann estimates ($HL$) of the median difference in absolute errors between REC-MILP-LW and integer-rounded base forecasts at the Market-Product Type level, and $\bar{d}$, the mean of these differences (Table \ref{tab: rec wilcoxon}); (iii) Paired Wilcoxon signed-rank test on the paired absolute error differences $d = AE_{\text{REC-MILP-LW}} - AE_{\text{Base (rounded)}}, d\neq 0$, with null hypothesis $H_0: \mathrm{median}(d) = 0$ and alternative $H_1: \mathrm{median}(d) < 0$ (i.e., REC-MILP-LW yields lower errors). We report the number of pairs ($N_r$), the Wilcoxon test statistic ($V$), the p-value ($p$), and the rank-biserial correlation ($r_{\text{rb}}$) as the effect size (Table \ref{tab: rec wilcoxon}). The tests were conducted using the Python \textit{SciPy} 1.16.1 library.

\begin{table}[h]
\centering
\begin{adjustbox}{width=0.9\textwidth,center=\textwidth}
\setlength\extrarowheight{-4pt}
\begin{tabular}{lllllll}
\toprule
Method & $N_r$ & $V$ & $p$ & $r_{\text{rb}}$ & $HL$ & $\bar{d}$ \\ \hline
ETS & 50 & 666 & 0.609 & 0.080 & 0.500 [-1.000 1.000] & 0.040 [-1.879 1.959] \\ 
ARIMA & 299 & 13346 & $3.425 \times 10^{-12} $ & -0.378 & -1.000 [-5.550 1.000] & -0.545 [-3.175 2.084] \\
LightGBM REC & 84 & 1312 & 0.017 & -0.333 & -1.000 [-230.725 97.325] & -8.000 [-120.153 104.153] \\
\bottomrule
\end{tabular}
\end{adjustbox}

\caption{Wilcoxon signed-rank test results comparing REC-MILP-LW and integer-rounded base forecasts at the Market–Product Type level.}\label{tab: rec wilcoxon}
\vspace{0.5em}
    \begin{minipage}{\textwidth}
        \footnotesize
        \textit{Notes.} Negative $HL$, $\bar{d}$, and $r_{\text{rb}}$ values indicate that REC-MILP-LW had lower absolute errors than the rounded base forecasts. \\
    \end{minipage}
\end{table}

\begin{table}[H]
\centering

\begin{adjustbox}{width=0.8\textwidth,center=\textwidth}
\setlength\extrarowheight{-4pt}

\begin{tabular}{lllllll}
\toprule
Method       & Level & Base & BU   & MinT & REC-MILP & REC-MILP-LW \\ \hline
ETS          & 0: Brand     & 0,313  & 0,632 & 0,433 & \textbf{0,318}    & \underline{\textbf{0,313}}       \\
             & 1: Market     & 0,338  & 1,194 & 0,513 & \underline{\textbf{0,337}}    & \textbf{0,338}       \\
             & 2: Market--Product Cluster     & 0,451  & 1,408 & 0,657 &\textbf{0,430}     & \underline{\textbf{0,429}}       \\
             & 3: Market--Product Line     & 0,656  & 1,000     & 0,686 & \textbf{0,657}    & \underline{\textbf{0,642}}       \\
             & 4: Market--Product Type     & 0,600    & 1,538 & \textbf{0,538} & \textbf{0,545}    & \underline{\textbf{0,529}}       \\
ARIMA        & 0: Brand     & 0,269  & 0,562 & 0,396 & \underline{\textbf{0,258}}    & 0,317       \\
             & 1: Market      & 0,304  & 0,712 & 0,393 & \underline{\textbf{0,309}}    & 0,327       \\
             & 2: Market--Product Cluster     & 0,454  & 0,888 & 0,562 & \underline{\textbf{0,411}}    & \textbf{0,415}       \\
             & 3: Market--Product Line     & 0,625  & 0,924 & 0,719 & 0,621    & \underline{\textbf{0,571}}       \\
             & 4: Market--Product Type     & 0,616  & 0,579 & 0,721 & 0,718    & \underline{\textbf{0,545}}       \\
LightGBM REC & 0: Brand     & 0,185  & 0,304 & 0,301 & \underline{\textbf{0,257}}    & 0,318       \\
             & 1: Market      & 0,350   & 0,359 & 0,352 & \underline{\textbf{0,324}}    & \textbf{0,333}       \\
             & 2: Market--Product Cluster     & 0,435  & \textbf{0,431} & \textbf{0,415} & \underline{\textbf{0,413}}    & 0,445       \\
             & 3: Market--Product Line     & 0,409  & 0,594 & 0,568 & \underline{\textbf{0,514}}    & 0,597       \\
             & 4: Market--Product Type     & 0,566  & \underline{\textbf{0,566}} & 0,666 & 0,619    & \underline{\textbf{0,566}}      \\
\bottomrule
\end{tabular}
\end{adjustbox}

\caption{Accuracy of reconciliation methods (WMAPE) across the hierarchy.}\label{tab: rec wmape}
\vspace{0.5em}
    \begin{minipage}{\textwidth}
        \footnotesize
        \textit{Notes.} Underlined values indicate the best accuracy, while bold numbers represent values that are at most 5\% worse than the best. The base forecast is not included in this comparison. 
    \end{minipage}
\end{table}

\begin{table}[H]
\centering

\begin{adjustbox}{width=0.8\textwidth,center=\textwidth}
\setlength\extrarowheight{-4pt}

\begin{tabular}{lllllll}
\toprule
Method       & Level & Base & BU   & MinT & REC-MILP & REC-MILP-LW \\ \hline
ETS          & 0: Brand     & 1,159  & 2,219 & 1,552 & \underline{\textbf{1,159}}    & \underline{\textbf{1,159}}       \\
             & 1: Market     & 0,644  & 1,741 & 0,898 & \underline{\textbf{0,647}}    & \textbf{0,650}        \\
             & 2: Market--Product Cluster     & 0,605  & 1,535 & 0,846 & \underline{\textbf{0,649}}    & \textbf{0,650}        \\
             & 3: Market--Product Line     & 0,566  & 1,042 & 0,660  & \textbf{0,632}    & \underline{\textbf{0,628}}       \\
             & 4: Market--Product Type     & 0,605  & \textbf{0,813} & \underline{\textbf{0,806}} & \textbf{0,823}    & \textbf{0,823}       \\
ARIMA        & 0: Brand     & 1,027  & 1,964 & 1,418 & \underline{\textbf{0,987}}    & 1,161       \\
             & 1: Market     & 0,628  & 1,253 & 0,710  & \underline{\textbf{0,648}}    & \textbf{0,657}       \\
             & 2: Market--Product Cluster     & 0,587  & 1,035 & 0,732 & 0,627    & \underline{\textbf{0,596}}       \\
             & 3: Market--Product Line      & 0,507  & 0,752 & \underline{\textbf{0,595}} & \textbf{0,607}    & \textbf{0,605}       \\
             & 4: Market--Product Type     & 0,609  & \underline{\textbf{0,764}} & 0,813 & 0,841    & \textbf{0,765}       \\
LightGBM REC & 0: Brand     & 0,764  & 1,141 & 1,154 & \underline{\textbf{1,006}}    & 1,086       \\
             & 1: Market     & 0,613  & \underline{\textbf{0,592}} & \textbf{0,608} & \textbf{0,617}    & 0,631       \\
             & 2: Market--Product Cluster     & 0,643  & 0,644 & \underline{\textbf{0,603}} & \textbf{0,606}    & 0,654       \\
             & 3: Market--Product Line      & 0,559  & \textbf{0,574} & \textbf{0,574} & \underline{\textbf{0,573}}    & 0,609       \\
             & 4: Market--Product Type     & 0,821  & \textbf{0,821} & \underline{\textbf{0,815}} & 0,866    & \textbf{0,837}    \\

\bottomrule
\end{tabular}
\end{adjustbox}

\caption{Accuracy of reconciliation methods (RMSSE) across the hierarchy.}\label{tab: rec rmsse}
\vspace{0.5em}
    \begin{minipage}{\textwidth}
        \footnotesize
        \textit{Notes.} Underlined values indicate the best accuracy, while bold numbers represent values that are at most 5\% worse than the best. The base forecast is not included in this comparison. 
    \end{minipage}
\end{table}

\newpage
\section{Mixed-effects model and statistical testing}\label{apx: mixed model}
To assess the impact of user-generated online information on forecast accuracy, we estimated a linear mixed-effects model of the form:

\begin{equation}
y_{ijk} = \beta_0 + \beta_1 \text{Online}_{ijk} + u_j + v_k + \epsilon_{ijk},   
\end{equation}
where $y_{ijk}$ denotes the forecast accuracy for time series $i$ in market $j$ and test period $k$; $\text{Online}_{ijk}$ is a binary indicator for the use of online data (1 if used, 0 otherwise); $\beta_0$ is the intercept representing the mean accuracy without online data; $\beta_1$ captures the fixed effect of including online data; $u_j \sim \mathcal{N}(0, \sigma^2_u)$ is a random effect accounting for unobserved market-level variation; $v_k \sim \mathcal{N}(0, \sigma^2_v)$ represents the random effect of test period differences; and $\epsilon_{ijk} \sim \mathcal{N}(0, \sigma^2)$ is the residual error term. Parameters were estimated using restricted maximum likelihood. 

As a robustness check, we conducted the Wilcoxon signed-rank test comparing paired forecast accuracy outcomes between models with and without online data. The hypotheses were defined as: $H_0: \mathrm{median}(d) = 0$ (no accuracy difference) and alternative $H_1: \mathrm{median}(d) < 0$ (online data reduces forecast errors), where $d=WMAPE_{\text{with online}} - WMAPE_{\text{without online}}, d \neq 0$. This test was applied separately for each forecasting approach and hierarchy level. We report two key statistical measures to aid interpretation.
First, the marginal $R^2$, representing the proportion of variance explained by fixed effects only. Second, the standardized effect ($r$), derived from the Wilcoxon signed rank test, representing the effect size of the forecast accuracy differences. Values of 0.1, 0.3, and 0.5 are commonly interpreted as small, moderate, and large effects, respectively.

The analysis was based on forecast errors (WMAPE) from the test periods described in Section \ref{sec: data}. Each forecasting model (LightGBM DIR, -REC, and DRFAM-PP)  was evaluated with and without user-generated online features across all markets and seven rolling forecast periods, covering all hierarchy levels. These errors served as input for both the linear mixed-effects model and the Wilcoxon signed-rank test. The tests were conducted using the Python \textit{SciPy} 1.16.1 library.

\section*{Declaration of Generative AI and AI-assisted technologies in the writing process}
During the preparation of this work, the authors used OpenAI’s ChatGPT in order to improve language clarity and readability. After using this tool, the authors reviewed and edited the content as needed and take full responsibility for the content of the publication.

\bibliographystyle{model5-names}
\bibliography{PhD_bib}

\newpage 

\clearpage
\thispagestyle{empty}
\vspace*{3cm}

\begin{center}
    {\Large Electronic companion}\\[1.0em]
    {\LARGE Automobile demand forecasting: Spatiotemporal and
hierarchical modeling, life cycle dynamics, and
user-generated online information}\\[2em]
    {\large Tom Nahrendorf, Stefan Minner, Helfried Binder, Richard Zinck}\\[2em]
\end{center}

\clearpage

\newpage
\renewcommand{\thetable}{EC.\arabic{table}}
\setcounter{table}{0}

\renewcommand{\thefigure}{EC.\arabic{figure}}
\setcounter{figure}{0}

\renewcommand{\thesection}{EC.\arabic{section}}
\setcounter{section}{0}

\newcommand*\emptycirc[1][1ex]{\tikz\draw (0,0) circle (#1);} 
\newcommand*\halfcirc[1][1ex]{%
  \begin{tikzpicture}
  \draw[fill] (0,0)-- (90:#1) arc (90:270:#1) -- cycle ;
  \draw (0,0) circle (#1);
  \end{tikzpicture}}
\newcommand*\fullcirc[1][1ex]{\tikz\fill (0,0) circle (#1);}

\newpage

\section{Computation details}
All computations were performed on a system with an Intel Core i7 processor (6 cores, 2.6 GHz) and 16 GB of RAM, running macOS Sonoma 14.7.3. The analysis was conducted in Python 3.11.10 using the following libraries: \textit{scikit-learn} 1.5.2 for direct (DIR) and hybrid (HYB) forecasting strategies and data pipelining; \textit{MLForecast} 1.0.2 (Nixtla) for recursive (REC) forecasting; \textit{NeuralForecast} 3.0.0 for deep learning models (DeepAR, NBEATSx); \textit{HierarchicalForecast} 1.2.1 for hierarchical reconciliation benchmarks; and the Gurobi Optimizer 12.0.1 for solving the proposed pooling selection model (POOL-SEL-BP), and mixed-integer linear programming reconciliation models (REC-MILP and REC-MILP-LW).

\section{Market clustering}\label{sec: market clustering}
Market clustering was introduced to reduce heterogeneity in demand dynamics across markets and to enable more homogeneous model training. Total market volumes were used in raw form, with no missing values detected. A total of 783 time-series features were extracted with the \textit{tsfresh} library (default settings), and all features were standardized prior to dimensionality reduction. Principal component analysis (\textit{scikit-learn}, default settings) reduced the feature space to 16 components, explaining 95\% of the variance. Standard k-means clustering (\textit{scikit-learn}, k-means++ initialization, objective: intra-cluster variance) was applied, with five clusters selected using the elbow method based on inertia and cluster size balance. This avoided degenerate solutions with many small clusters. The resulting clusters comprised 2, 9, 1, 1, and 10 markets, respectively, covering all 23 markets.

For robustness, we evaluated clustering based on dynamic time warping (DTW). However, DTW-based clustering resulted in higher forecast bias in the test period (Year 12). Using a LightGBM model trained separately per cluster, the forecast bias was 6,056 for feature-based clustering, 11,501 for DTW clustering, and 11,510 for a global model without clustering.

\section{Pooling set selection}
Table \ref{tab: pools} summarizes the candidate pooling families $\mathcal{G}$ considered in POOL-SEL-BP. Each pool $g$ represents a feasible partition of training data; the reported quantities describe its structural and cost characteristics. The pool size $M_g$ denotes the number of models trained under that grouping, and $G_g$ is the Gini coefficient of the within-pool series sample-share distribution (higher = more imbalance). The no-pooling baseline, representing local models trained individually per series (naive forecast), was included with $\kappa_{\text{no pooling}}=0$ to provide a zero-cost fallback and guarantee feasibility. In total, ten candidate pools plus no-pooling were evaluated (hierarchical, domain-based, clustering-based, and combined forms). Other candidate combinations (e.g., Market–Product Line, implying 23 × 17 = 391 models for each direct and recursive learner) were considered but deemed non-feasible by the industry partner due to excessive model-training complexity.

\begin{table}[H]
    \centering
    \resizebox{0.8\textwidth}{!}{
    \setlength\extrarowheight{-6pt}
    \begin{tabular}{llll}
    \toprule
      Pool $g$  & Pool size ($M_g$) & Sample imbalance ($G_g$) & Pool cost ($\kappa_g$) \\ \hline
     \textit{Hierarchical} & & & \\
    Market & 23 & 0.1267 &  0.0148 \\
    Product Cluster & 5 & 0.1807 &  0.0092 \\
    Product Line & 14 & 0.4666 & 0.0243 \\
    Market--Product Cluster & 92 & 0.2409 &  0.0487 \\
    & & &\\
    \textit{Domain-based} & & & \\
    Market maturity & 2 & 0.1900 &  0.0083 \\
    Fuel type & 3 & 0.6346 &  0.0262 \\
    Body type & 8 & 0.3115 & 0.0156 \\
    & & &\\
    \textit{Clustering-based} & & & \\
    Market cluster\textsuperscript{*} & 5 & 0.4146 &  0.0184 \\
    & & &\\
    \textit{Combinations} & & & \\
    Market cluster--fuel type & 15 & 0.7753 & 0.0368 \\
    Market cluster--body type & 40 & 0.5255 & 0.0377 \\
   
      \bottomrule
    \end{tabular}
    }
    \caption{Candidate pooling families and associated costs.}
    \label{tab: pools}
    \vspace{0.5em}
    \begin{minipage}{\textwidth}
        \footnotesize
        \textit{Notes. \textsuperscript{*}: For specification of the market clustering procedure, see e-companion \ref{sec: market clustering}}.
    \end{minipage}
\end{table}

To capture both computational complexity and risk of overfitting, we defined the pool cost as $\kappa_g = s M_g(1+\nu G_g)$, where the scale factor $s$ aligns the cost magnitude with the average validation loss and $\nu\geq0$  controls the penalty for imbalance. We set $\nu = 1$ for an equal trade-off between model count and imbalance and chose $s \approx 0.0002$ so that the mean of $\kappa_g$ matched the mean relative loss, giving $\lambda$ an interpretable accuracy–complexity trade-off (a one-unit change in $\lambda$ roughly trades one average series-level improvement against opening one additional pool). The optimization used relative losses $\Delta L_{ig}= L_{ig} - L_{i0}$ based on RMSSE, where $L_{ig}$ is the per-fold mean of direct and recursive model losses, subsequently averaged across 12 rolling validation folds; $L_{i0}$ denotes the corresponding no-pooling baseline.

\begin{figure}[H]
  \centering
  \begin{minipage}{\textwidth}
    \centering
    \includegraphics[width=0.9\textwidth]{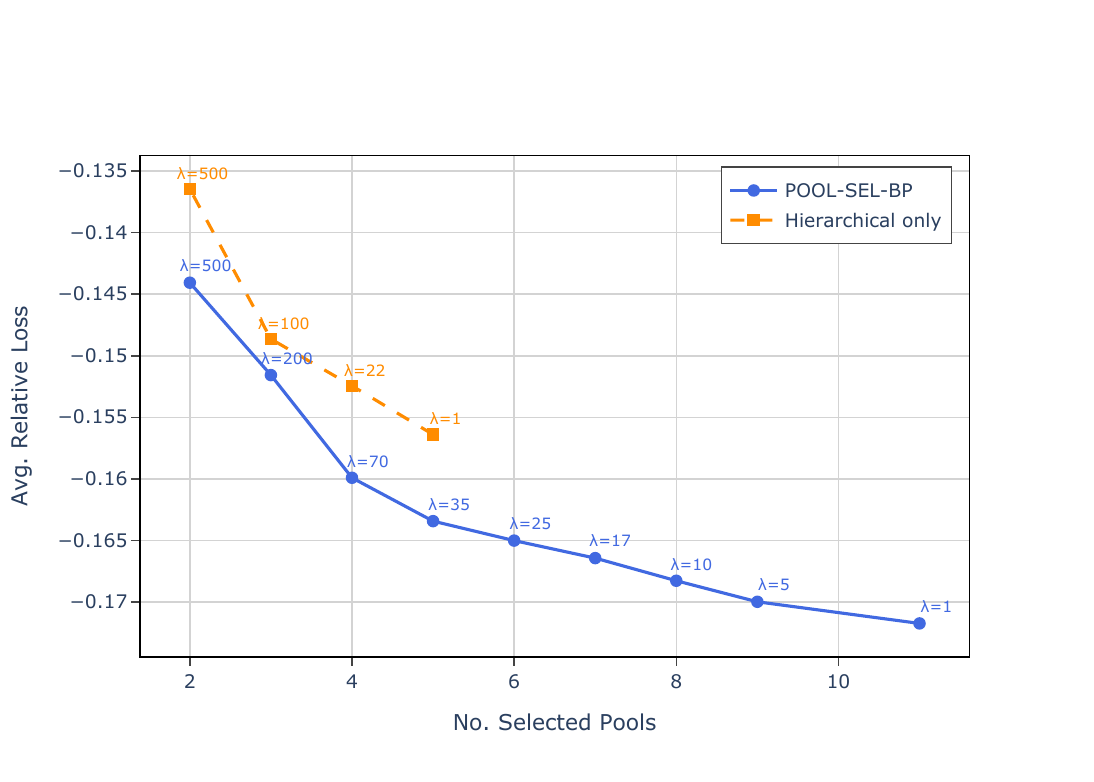}
    \caption{POOL-SEL-BP hyperparameter tuning.}  \label{fig: lambda tuning}
  \end{minipage}
\end{figure}

To calibrate the trade-off parameter $\lambda$, we solved POOL-SEL-BP for a grid of ten candidate values and recorded both the average relative validation loss and the number of pools activated. Figure \ref{fig: lambda tuning} shows the resulting accuracy–complexity frontier. Each point represents the mean relative RMSSE (improvement over the local baseline) achieved for a given $\lambda$, plotted against the corresponding number of open pools. The curve is monotonically decreasing with a clear elbow at $\lambda=35$, beyond which additional pools yield diminishing returns. We therefore set $\lambda=35$, which opens five pools. At comparable model budgets, optimal pooling under POOL-SEL-BP outperforms fixed hierarchical pooling, confirming gains from endogenously selecting pooling structures based on validation performance.

Table \ref{tab: results pool selection} reports the optimal solution of POOL-SEL-BP. It identified four dominant pooling structures based on their cumulative improvement in relative RMSSE over the no-pooling baseline. For subsequent forecasting experiments, we selected the top three pools (body type, market, and market cluster) as input configurations for DRFAM-PP.

\begin{table}[H]
    \centering
    \resizebox{0.7\textwidth}{!}{
    \setlength\extrarowheight{-6pt}
    \begin{tabular}{llll}
    \toprule
      Pool $g$  & $n$ assigned & Sum of rel. pool losses & Avg. rel. pool loss \\ \hline
    Body type & 114 & -37.1718 & -0.3260 \\
    Market & 64 & -20.3019 & -0.3172 \\
    Market cluster & 46 & -18.4399 & -0.4008 \\
    Market maturity & 91 & -17.8540 & -0.1962 \\
      \bottomrule
    \end{tabular}
    }
    \caption{Results pool selection optimization.}
    \label{tab: results pool selection}
\end{table}

Body-type pooling captures structural demand heterogeneity across vehicle segments, which reflect distinct usage patterns, life cycle dynamics, and price positioning (e.g., compact vs. SUV). Market-level pooling leverages country-specific macro and regulatory conditions that shape local sales trajectories. Market-cluster pooling extends this logic by grouping markets with similar seasonal behavior, trend, and volatility patterns, enabling transfer of information across comparable regions.

\section{Geometry of reconciliation}
\begin{figure}[H]
\centering
\includegraphics[scale=0.7]{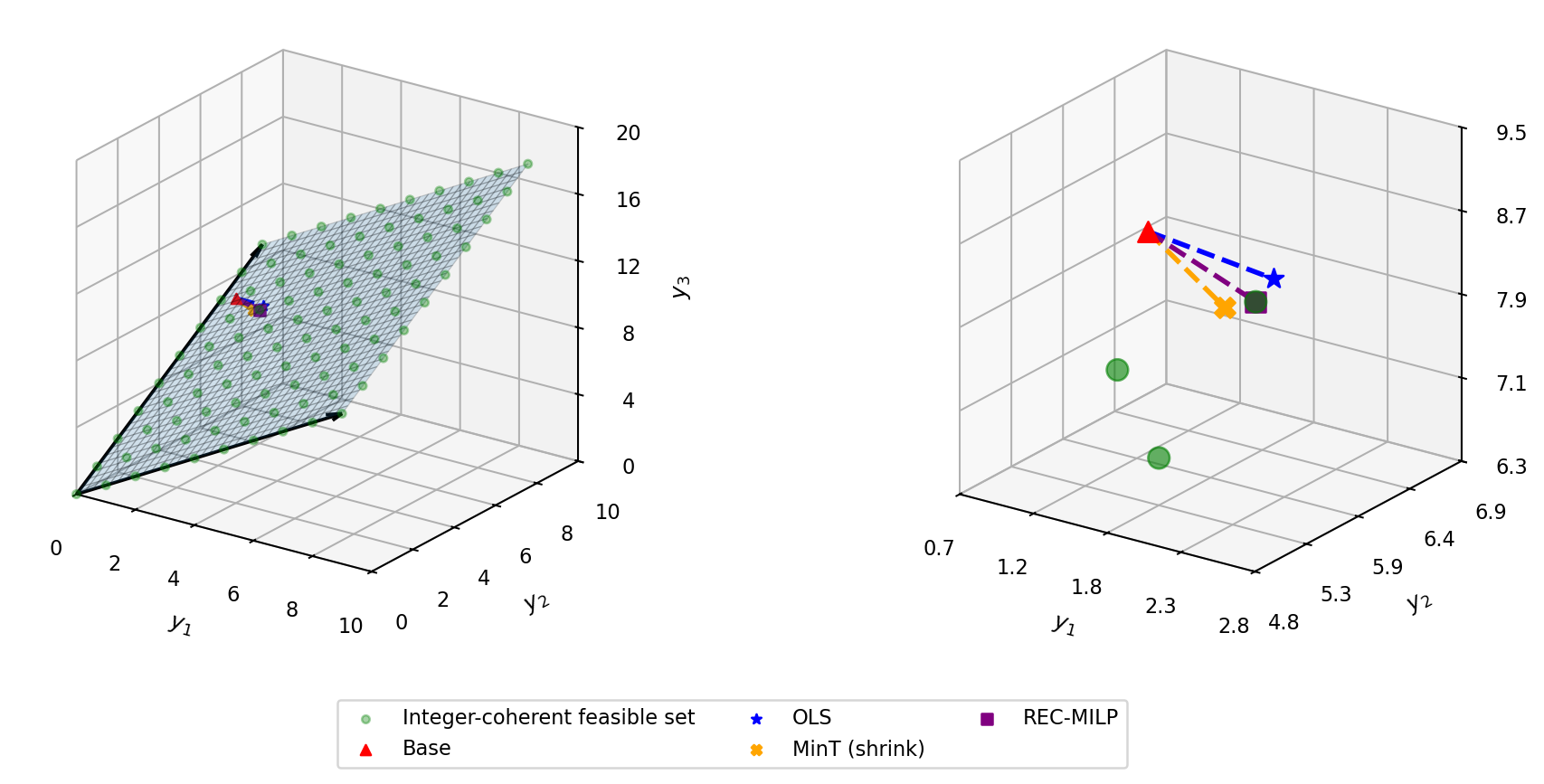}
\caption{Geometry of reconciliation.}\label{fig: geometry reconciliation}
\end{figure}

We illustrate reconciliation in the minimal three–series hierarchy with two bottom series and their aggregate. Let $y_1$ and $y_2$ denote bottom–level series and $y_3$ the aggregate, subject to the coherence constraint $y_3 = y_1 + y_2$. In Figure \ref{fig: geometry reconciliation}, the coherent subspace $\mathcal{C}=\{\, a\,\mathbf{v}_1 + b\,\mathbf{v}_2 \; : \; a,b\in[0,1] \}$ appears as a parallelogram spanned by $\mathbf{v}_1=(9,0,9)^{\top}$ and $\mathbf{v}_2=(0,9,9)^{\top}$. We take the base (incoherent) forecast $\mathbf{P}=(1.5, 5.6, 8.7)^{\top}$, so $y_1+y_2=7.1\neq y_3$. Each reconciler maps $\mathbf{P}$ to the closest coherent point in $\mathcal{C}$, where "closest" is defined by its metric:

\begin{enumerate}
    \item OLS (identity metric, $L_2$-norm): orthogonal (Euclidean) projection onto $\mathcal{C}$ gives $\mathbf{\tilde{y}}_{\text{OLS}} = (2.033,6.133,8.167)^{\top}$.
    \item MinT (weighted, $L_2$-norm): projection under metric $\mathbf{W}$. We use MinT–shrink constructed from an illustrative forecast–error covariance
  $
  \mathbf{\Sigma}=\begin{bmatrix}
  4.0 & 1.8 & 5.8 \\
  1.8 & 9.0 & 10.8 \\
  5.8 & 10.8 & 16.6
  \end{bmatrix},
  \mathbf{W}=\big((1-\lambda)\mathbf{\Sigma}+\lambda\,\mathrm{diag}(\mathbf{\Sigma})\big)^{-1}, \lambda=0.3,
  $
  yielding $\tilde{\mathbf{y}}_{\text{MinT}}=(1.716,\,6.086,\,7.803)^{\top}$. 

    \item REC-MILP (integer, $L_1$-norm): $L_1$ projection restricted to the integer–coherent lattice
  $\mathcal{C}\cap\mathbb{Z}^3$ (green points) gives
  $\tilde{\mathbf{y}}_{\text{REC–MILP}}=(2,\,6,\,8)^{\top}$.
\end{enumerate}

\section{Minor empirical results}
\subsection{No single forecasting method is optimal across hierarchy levels}
Our results confirm that no single forecasting method dominates at all hierarchical levels. Demand dynamics differ substantially between strategic levels (e.g., brand or market) and operational levels (e.g., product type). This implies that level-specific forecasting strategies are necessary to match the volatility, data richness, and planning horizon of each level. For automotive original equipment manufacturers (OEMs), this supports the adoption of hierarchical modeling frameworks, where methods are tailored to both aggregated and disaggregated decision contexts.

\subsection{Forecast accuracy decreases with increasing granularity}
Forecast accuracy, measured in weighted mean absolute percentage error (WMAPE), decreases as the level of granularity increases, as WMAPE is directly comparable across hierarchical levels, whereas root mean squared scaled error (RMSSE) is scale-dependent and not invariant across aggregation levels. Aggregated levels (brand, market) achieve the highest accuracy, while forecasts at the product line and product type levels show higher volatility (Table \ref{app: accuracy table}). This finding highlights the need for OEMs to anticipate higher uncertainty in granular demand forecasts and to incorporate probabilistic forecasts into operational planning, particularly for production and inventory allocation at the product type level.

\subsection{Hybrid direct–recursive forecasting improves bottom-level accuracy}
At the most granular level (Market--Product Type), HYB outperforms pure DIR and REC approaches as well as deep
learning models by combining DIR and REC methods
through chained forecasting (Table \ref{app: accuracy table}). This approach leverages the horizon-specific accuracy of DIR
forecasts alongside the temporal consistency of REC models, which incorporate information
from previous predictions to maintain coherence across time steps. While HYB strategies did not
outperform at higher levels of the hierarchy, they proved especially effective for operational-level
forecasting by delivering stable and consistent planning inputs.

\subsection{Probabilistic forecasting at the bottom level}
\begin{figure}[H]
    \centering
    \includegraphics[scale=0.6]{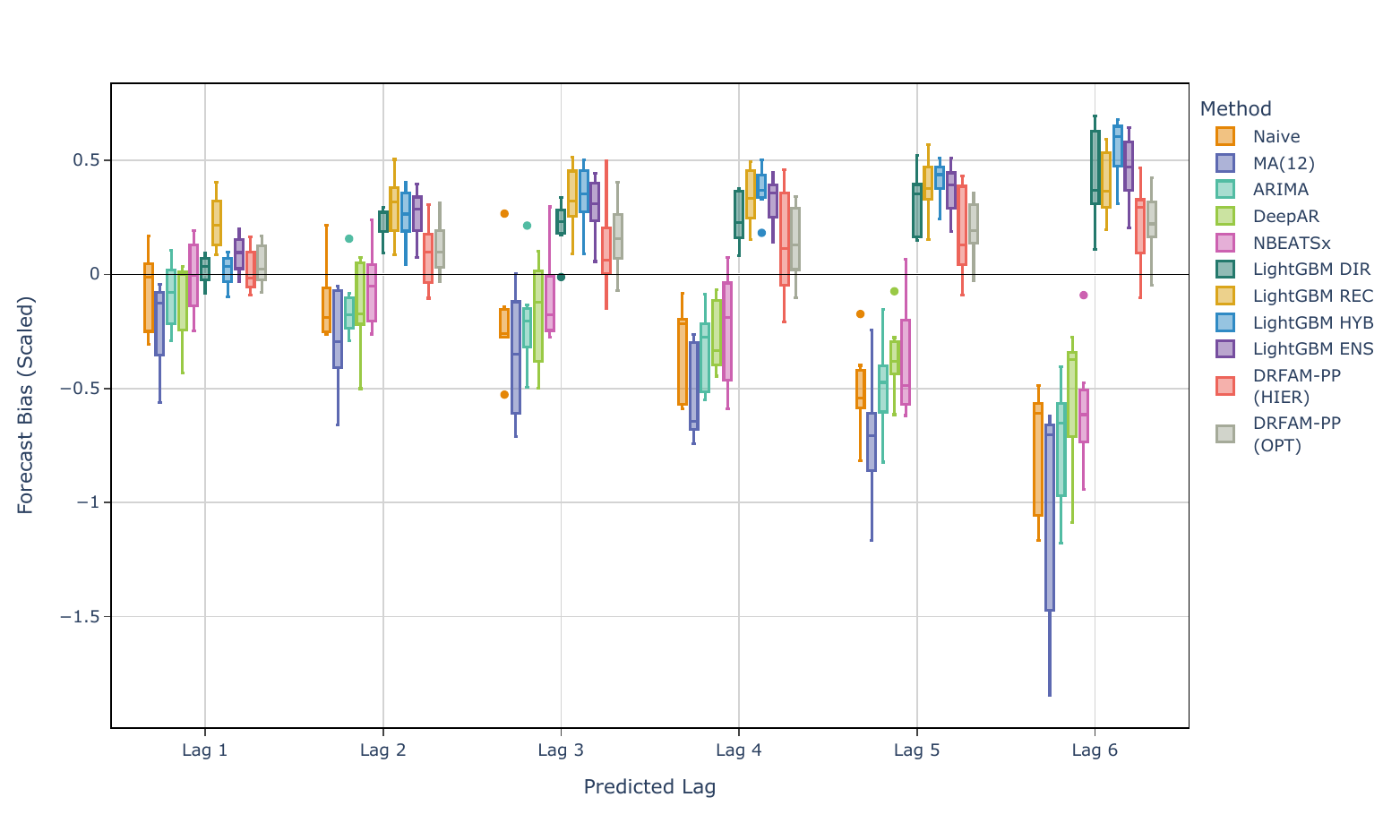}
    \caption{Forecast bias}\label{fig: fb bias}
\end{figure}

\begin{figure}[H]
    \centering
    \includegraphics[scale=0.6]{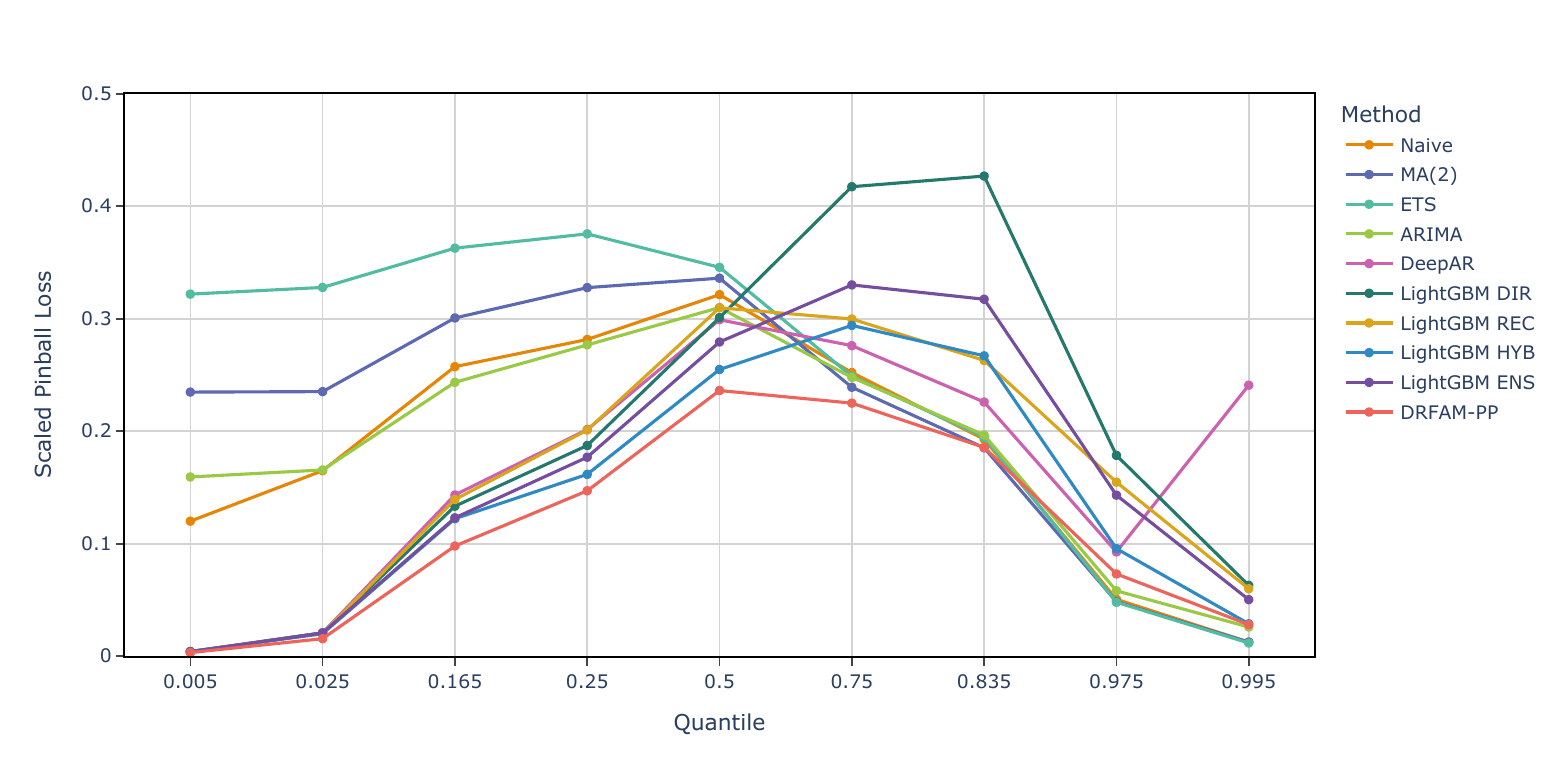}
    \caption{Scaled pinball loss}\label{fig: pinball loss}
\end{figure}

At the lowest hierarchy level, the Direct Recursive Forecast Averaging Method via Partial Pooling (DRFAM-PP) consistently outperforms other models across most quantiles (see \ref{sec: pinball loss}, Figure \ref{fig: pinball loss}), delivering robust probabilistic forecasts. Its performance is strongest at lower and mid
quantiles, while at the upper quantiles (e.g., $q = 0.975, 0.995$), simple statistical benchmarks
such as Naive and ARIMA slightly outperform it due to their inherent overestimation bias (see \ref{sec: forecast bias}, Figure \ref{fig: fb bias}). In contrast, tree-based global models (LightGBM DIR, -REC, and -HYB) perform well at lower quantiles but show declining accuracy for quantiles above $q \geq 0.75$, particularly when using DIR. This underperformance is largely attributed to the lack of temporal smoothing and the inability to capture sequential dependencies
in DIR models. For instance, DeepAR maintains better performance in these high quantiles by leveraging recurrent
structures to model time-dependent patterns.

\subsection{Pricing features as locally influential predictors}

\begin{figure}[htpb]
  \centering
  \begin{minipage}{\textwidth}
    \centering
    \includegraphics[width=0.9\textwidth]{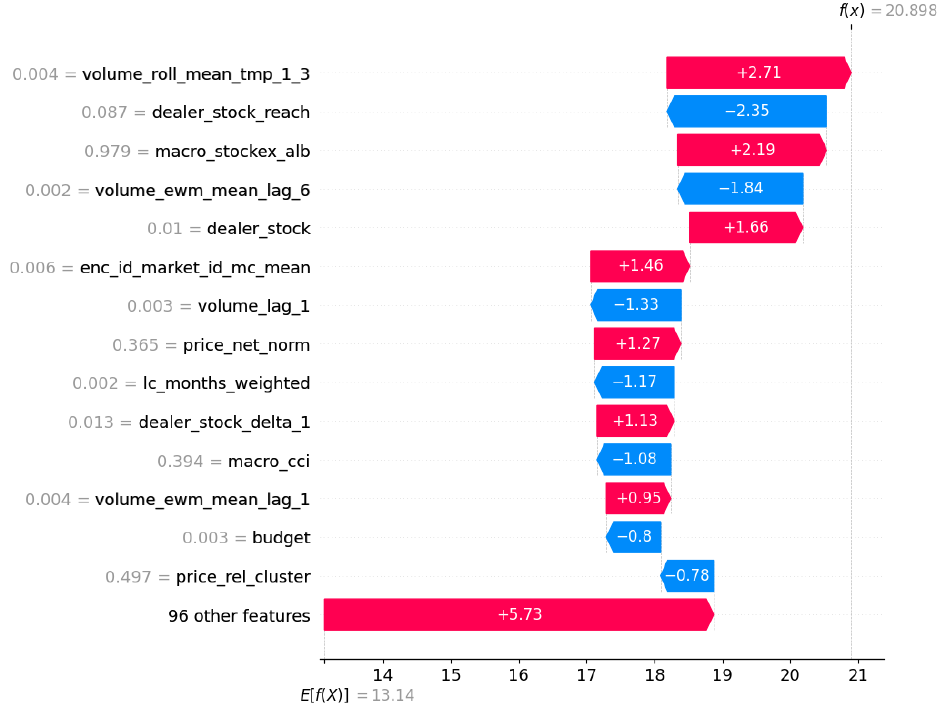}
    \caption{SHAP waterfall plot for forecast step 2, providing a local explanation for a specific market-product type combination.}  \label{fig:shap waterfall}

  \end{minipage}
\end{figure}

Although price-related features did not surface prominently in the global SHAP analysis, local interpretability plots (Figure \ref{fig:shap waterfall}) show that they can strongly affect individual predictions. Aligned with economic theory, lower product prices (price\_net\_norm) increase predicted demand. However, a low relative price compared to the series cluster average (price\_rel\_cluster) can lead to lower
predicted demand. This reflects perceived value dynamics: in premium markets, products priced well below peers signal lower quality, decreasing demand.

\newpage

\begin{landscape}
\subsection{Point and probabilistic forecast accuracy across hierarchical levels}\label{app: accuracy table}
\setlength{\tabcolsep}{2pt}  
\setlength\extrarowheight{0pt}

\begin{table}[H]
\centering{ 
\begin{adjustbox}{width=1.33\textwidth,center=\textwidth}
\setlength\extrarowheight{-2pt}

\begin{tabular}{lllllllllllllllllllllllll}

\toprule 
Model & \multicolumn{6}{l}{RMSSE}           & \multicolumn{6}{l}{WMAPE [x 100\%]}           & \multicolumn{6}{l}{Forecast bias (scaled)}   &  \multicolumn{6}{l}{MSPL}              \\ \cline{2-25} 
      & \multicolumn{5}{l}{Level ID} & Avg. & \multicolumn{5}{l}{Level ID} & Avg. & \multicolumn{5}{l}{Level ID} & Avg. & \multicolumn{5}{l}{Level ID} & Avg.\\ \cline{2-6} \cline{8-12} \cline{14-18} \cline{20-24} 
      & 0    & 1   & 2   & 3   & 4   &      & 0    & 1   & 2   & 3   & 4   &      & 0    & 1   & 2   & 3   & 4   & &  0 & 1& 2& 3 &4 &  \\ \hline
 \textit{Statistical} &      &     &     &     &     &      &      &     &     &     &     &      &      &     &     &     &     &  & & & & &  \\  
Naive &   \underline{\textbf{0.640}}   &  0.650   &  0.629   &  0.644   &  0.547   &     \textbf{0.622} &  \textbf{0.156}    &  0.351   &   0.545  &  0.746   &   1.265  &    0.613  & \underline{\textbf {-0.103}}	&  -0.089   &  -0.092   &  -0.385   &  -0.241   &  -0.182 & 0.361 &0.372& 0.406& 0.374& 0.321 & 0.367\\  
sNaive &  1.029    &  0.680   &  0.828   &   0.857  &  0.862   &    0.851  &  0.265    &   0.397  &  0.711   &   0.963  &  1.082   &   0.684   &   -0.265   &   -0.273  &  -0.347   &  -0.581   &  -0.592   &  -0.412 & 0.612& 0.400& 0.487& 0.490& 0.460 &0.490\\  
MA(2) &  \textbf{0.652}    &  0.687   &  0.609   &  0.621   & 0.563    &     \textbf{0.626} &  \underline{\textbf{0.149}}    &   0.382  &  0.543   &  0.722   &  0.847   &    0.529  &   -0.109   & -0.167    &  -0.130   &  -0.464   & -0.317    &  -0.237 & 0.334& 0.379 &0.391 &0.377& 0.336& 0.363\\  
MA(3) &  \textbf{0.664}    &   0.648  &  0.609   &  0.579   &  0.559   &    \underline{\textbf{0.612}}  & 0.165     &  0.376   &   0.518  &  0.667   &   0.868  &    0.519  &   -0.115   &  -0.132   &  -0.154   &  -0.358   &  -0.343   &  -0.220 & 0.369& 0.369& 0.418 &0.373& 0.341& 0.374\\  
MA(6) &   0.832   &  0.710   &   0.673  &  0.632   &   0.614  &     0.692 &    0.201  &   0.362  &  0.539   &  0.764   &  0.961   &    0.565  &  	-0.183    & -0.226    & -0.249    & -0.543    &  -0.500   &  -0.340 & 0.465& 0.432& 0.423& 0.400& 0.477& 0.419 \\  
MA(12) &  0.950    &   0.658  &   0.663  &  0.657   &  0.606   &    0.707  &  0.229    & 0.343    &  0.538   &   0.771  &  1.056   &  0.587    &  -0.224    &  -0.244   &  -0.272   &  -0.673   &  -0.781   &  -0.439 & 0.529 &0.380& 0.433 &0.407 &0.388& 0.427 \\  
ETS &   0.766   &  0.670   &   0.608  &  0.616   &  0.577   &      0.647&  0.185    &  0.365   &   0.506  &  0.746   &  1.265   &      0.613&  -0.158	    &   -0.209  &   -0.194  &  -0.581   & -0.976    &  -0.423 & 0.427 &0.362 &0.406 &0.362& 0.346& 0.381\\  
ARIMA &   0.982   &  0.685   &   0.631  &  0.579   &  0.539   &     0.683 &  0.250    &  0.384   &   0.498  &  0.709   &  1.000   &    0.568  &  -0.235    &  -0.235   & -0.176    &  -0.375   &  -0.311   &  -0.313 & 0.560 &0.423& 0.390 &0.535& 0.310 &0.407\\  
Prophet &   1.199   &  0.743   &  0.785   &  0.773   &  0.949   &   0.890   &  0.304    &  0.497   &  0.650   &   1.055  &  1.797   &   0.861   &  -0.304	    &  -0.385   &  -0.383   &  -0.942   &  -1.520   & 0.703 & 0.695& 0.465 &0.485& 0.455& 0.549& 0.530 \\  
 &      &     &     &     &     &      &      &     &     &     &     &      &      &     &     &     &     &   \\  
\textit{ML-based} &      &     &     &     &     &      &      &     &     &     &     &      &      &     &     &     &     &   \\  
DeepAR &   \dag   &   \dag  &  \dag   &  \dag   &  0.494   &   \dag   &   \dag  &    \dag &  \dag   &   \dag  &  0.935   &   \dag   &   \dag   &  \dag   &   \dag  &   \dag  &   -0.378  &  \dag & \dag &\dag &\dag&\dag &0.182& \dag\\  
NBEATSx &    \dag  &  \dag   &  \dag   &   \dag  &   0.531  &    \dag  &   \dag   &  \dag   &  \dag   &  \dag   &  1.000   &   \dag   &  \dag    &  \dag   &   \dag  &  \dag   &  -0.429   & \dag  & \dag& \dag &\dag& \dag& 0.654& \dag\\  
LightGBM DIR & 1.145     & \textbf{0.606}    &  0.621   &  0.521   & 0.581 &   0.695&  0.246    &  0.347    &  0.496   &  0.639   &  0.731   &    0.492 &   
-0.257    & -0.117    &  \underline{\textbf{-0.053}}   &   -0.226   &  -0.192   & -0.169 & 0.320 &0.249 &0.229 &0.185 &0.227 &0.242\\
LightGBM REC &  0.750    &   0.638  &  0.618   &  0.543   &  0.576   &  \textbf{0.625}    &  0.170    &  0.350   &  0.504   &  0.600   &  0.742   &   0.473   &  -0.134    &  -0.265   &  -0.108   &    -0.304   &  -0.179   & -0.198 & \underline{\textbf{0.233}} &\underline{\textbf{0.202}} &\textbf{0.206} &0.155& 0.171 &\underline{\textbf{0.193}} \\  
LightGBM HYB & 1.267     &  0.730   &  0.804   &  0.675   &   0.478  &  0.791   &  0.306    &  0.387   &  0.700   & 0.835    &  \textbf{0.677}   &   0.581   &  -0.300    &  \underline{\textbf{0.041}}   &   0.649   &  0.822   &  0.469   &  0.336 & 0.292& 0.217 &0.257& 0.206& 0.154& 0.225\\  
LightGBM ENS &  1.020    &  \underline{\textbf{0.579}}   &  \underline{\textbf{0.546}}   &  \underline{\textbf{0.436}}   &  0.516   &   \textbf{0.619}   &  0.250    &  \underline{\textbf{0.303}}   &  \underline{\textbf{0.471}}   &  \underline{\textbf{0.518}}   &  \underline{\textbf{0.654}}   &  \underline{\textbf{0.439}}    &   -0.229    & -0.094    &  0.143    &  \underline{\textbf{0.086}}   &  \underline{\textbf{0.026}}   &  \underline{\textbf{-0.014}} & \textbf{0.240} &0.206& \underline{\textbf{0.200}}& \underline{\textbf{0.145}}& 0.181 &\textbf{0.194}\\  
DRFAM-PP (HIER) &  \dag    &   \dag  &  \dag   &  \dag   &  0.469   &   \dag   &   \dag   & \dag    &  \dag   &  \dag   &  \underline{\textbf{0.654}}   &  \dag    &    \dag  &    \dag &   \dag  &  \dag  &   0.348   &  \dag & \dag& \dag &\dag &\dag & \textbf{0.134} &\dag \\  
DRFAM-PP (OPT) &  \dag    &   \dag  &  \dag   &  \dag   &  \underline{\textbf{0.418}}   &   \dag   &   \dag   & \dag    &  \dag   &  \dag   &  0.695   &  \dag    &    \dag  &    \dag &   \dag  &  \dag  &   0.500   &  \dag & \dag& \dag &\dag &\dag & \underline{\textbf{0.128}} &\dag \\

\bottomrule

\end{tabular}
\end{adjustbox}
}

\caption{Accuracy of the base forecast over seven test periods by benchmark model and level.}\label{Tab: wrmsse base}

\vspace{0.5em}
    \begin{minipage}{\textwidth}
        \footnotesize
        \textit{Notes.} \dag: Insufficient data available for the application of the specified model. Underlined values indicate the best accuracy, while bold numbers represent values that are at most 5\% worse than the best. 
    \end{minipage}

\end{table}
\end{landscape}

\section{Additional plots and tables}
\subsection{Macroeconomic and financial data}

\begin{table}[H]
    \centering
    \resizebox{\textwidth}{!}{
    \setlength\extrarowheight{-6pt}
    \begin{tabular}{lll}
    \toprule
      Indicator   & Unit & Data source\\ \hline
      \textbf{Macroeconomic} & & \\
      Unemployment rate   & \%  & Eurostat: ei\_lmhr\_m\\ 
      Inflation rate & \% & Eurostat: prc\_hicp\_manr \\
    Interest rate & \% & Eurostat: irt\_lt\_mcby\_m \\
    Exchange rate & National currency/ EUR & Eurostat: ert\_bil\_eur\_m \\
    GDP & Million EUR & Eurostat: namq\_10\_gdp \\
    \textbf{Harmonized consumer price index} & & \\
    All items & Index (2015 = 100)& Eurostat: prc\_hicp\_midx/CP00 \\
    New vehicles & Index (2015 = 100)& Eurostat: prc\_hicp\_midx/CP07111 \\
    Personal transport equipment -- Parts & Index (2015 = 100)& Eurostat: prc\_hicp\_midx/CP0721 \\
    Personal transport equipment -- Maintenance & Index (2015 = 100)& Eurostat: prc\_hicp\_midx/CP0723 \\
    Personal transport equipment -- Other services & Index (2015 = 100)& Eurostat: prc\_hicp\_midx/CP0724 \\
    Personal transport insurance & Index (2015 = 100)& Eurostat: prc\_hicp\_midx/CP1254 \\
    Public transport & Index (2015 = 100)& Eurostat: prc\_hicp\_midx/CP073 \\
    Electricity & Index (2015 = 100)& Eurostat: prc\_hicp\_midx/CP0451 \\
    Diesel & Index (2015 = 100)& Eurostat: prc\_hicp\_midx/CP07221 \\
    Petrol & Index (2015 = 100)& Eurostat: prc\_hicp\_midx/CP07222 \\
\textbf{Financial market} & & \\ 
\textit{Automotive manufacturers} & & \\
BMW AG & EUR & Yahoo Finance: BMW.DE \\
Mercedes-Benz Group AG & EUR & Yahoo Finance: MBG.DE \\
Volkswagen AG & EUR & Yahoo Finance: VOW3.DE\\
Tesla Inc. & USD & Yahoo Finance: TSLA\\
Toyota Motor Corporation & USD & Yahoo Finance: TM\\
Nissan Motor Co., Ltd. & USD & Yahoo Finance: NSANY\\
Honda Motor Co., Ltd. & USD & Yahoo Finance: HMC\\
Ford Motor Company & USD & Yahoo Finance: F\\
Stellantis N.V. & USD & Yahoo Finance: STLA\\
Mazda Motor Corporation & USD & Yahoo Finance: MZDAY \\
\textit{Supplier} & & \\
Continental AG (Parts) & EUR & Yahoo Finance: CON.DE \\
Aptiv PLC (Electronics) & USD & Yahoo Finance: APTV \\
Alcoa Corporation (Aluminium) & USD & Yahoo Finance: AA \\
Nucor Corporation (Steel)& USD & Yahoo Finance: NUE \\
Albemarle Corporation (Batteries) & USD & Yahoo Finance: ALB \\
QUALCOMM Incorporated (Semiconductors) & USD & Yahoo Finance: QCOM \\

      \bottomrule
    \end{tabular}
    }
    \caption{Macroeconomic and financial indicators used as explanatory variables.}
    \label{tab: macro data}
    \vspace{0.5em}
    \begin{minipage}{\textwidth}
        \footnotesize
        \textit{Notes.} The data extraction period spans 12 years, from Year 1 to Year 12, where available. Data were collected for 23 markets, but country names are withheld due to confidentiality agreements. All data are reported at monthly frequency, except for gross domestic product (GDP), which is published quarterly. 
    \end{minipage}
\end{table}

\subsection{Selected features by hierarchy level}

\begin{table}[H]
    \centering
    \resizebox{\textwidth}{!}{
    \setlength\extrarowheight{-6pt}
    \begin{tabular}{llllll}
    \toprule

Feature & \multicolumn{5}{l}{Level ID} \\
\cline{2-6} & 0 & 1 & 2 & 3 & 4 \\ \hline
\textbf{Demand} & & & & & \\ 
  Lags 1-12  & \emptycirc[0.5ex] \halfcirc[0.5ex] \fullcirc[0.5ex]&\emptycirc[0.5ex] \halfcirc[0.5ex] \fullcirc[0.5ex] &\emptycirc[0.5ex] \halfcirc[0.5ex] \fullcirc[0.5ex] &\emptycirc[0.5ex] \halfcirc[0.5ex] \fullcirc[0.5ex] & \emptycirc[0.5ex] \halfcirc[0.5ex] \fullcirc[0.5ex]\\

  Rolling minimum/maximum (windows: 1,3,6)  & \emptycirc[0.5ex] \halfcirc[0.5ex] \fullcirc[0.5ex]&\emptycirc[0.5ex] \halfcirc[0.5ex] \fullcirc[0.5ex] &\emptycirc[0.5ex] \halfcirc[0.5ex] \fullcirc[0.5ex] &\emptycirc[0.5ex] \halfcirc[0.5ex] \fullcirc[0.5ex] & \emptycirc[0.5ex] \halfcirc[0.5ex] \fullcirc[0.5ex]\\

  Rolling sum (windows: 2,3,12)  & \emptycirc[0.5ex] \halfcirc[0.5ex] \fullcirc[0.5ex]&\emptycirc[0.5ex] \halfcirc[0.5ex] \fullcirc[0.5ex] &\emptycirc[0.5ex] \halfcirc[0.5ex] \fullcirc[0.5ex] &\emptycirc[0.5ex] \halfcirc[0.5ex] \fullcirc[0.5ex] & \emptycirc[0.5ex] \halfcirc[0.5ex] \fullcirc[0.5ex]\\

  Rolling standard deviation (windows: 3)  & \emptycirc[0.5ex] \halfcirc[0.5ex] \fullcirc[0.5ex]&\emptycirc[0.5ex] \halfcirc[0.5ex] \fullcirc[0.5ex] &\emptycirc[0.5ex] \halfcirc[0.5ex] \fullcirc[0.5ex] &\emptycirc[0.5ex] \halfcirc[0.5ex] \fullcirc[0.5ex] & \emptycirc[0.5ex] \halfcirc[0.5ex] \fullcirc[0.5ex]\\

  Shifted rolling mean (windows: 3, shifts: 1,2,3,4)  & \emptycirc[0.5ex] \halfcirc[0.5ex] \fullcirc[0.5ex]&\emptycirc[0.5ex] \halfcirc[0.5ex] \fullcirc[0.5ex] &\emptycirc[0.5ex] \halfcirc[0.5ex] \fullcirc[0.5ex] &\emptycirc[0.5ex] \halfcirc[0.5ex] \fullcirc[0.5ex] & \emptycirc[0.5ex] \halfcirc[0.5ex] \fullcirc[0.5ex]\\

   Exponential weighted smoothing (lags: 1,6,12)  & \emptycirc[0.5ex] \halfcirc[0.5ex] \fullcirc[0.5ex]&\emptycirc[0.5ex] \halfcirc[0.5ex] \fullcirc[0.5ex] &\emptycirc[0.5ex] \halfcirc[0.5ex] \fullcirc[0.5ex] &\emptycirc[0.5ex] \halfcirc[0.5ex] \fullcirc[0.5ex] & \emptycirc[0.5ex] \halfcirc[0.5ex] \fullcirc[0.5ex]\\

   Historic mean/standard deviation of demand by market-product type  & & & & & \emptycirc[0.5ex] \halfcirc[0.5ex] \fullcirc[0.5ex]\\

   Historic mean/standard deviation of demand by product type  & & & & & \emptycirc[0.5ex] \halfcirc[0.5ex] \fullcirc[0.5ex]\\

\textbf{Temporal} & & & & &\\
 Binary encoding of year/quarter  & \emptycirc[0.5ex] \halfcirc[0.5ex] \fullcirc[0.5ex]&\emptycirc[0.5ex] \halfcirc[0.5ex] \fullcirc[0.5ex] &\emptycirc[0.5ex] \halfcirc[0.5ex] \fullcirc[0.5ex] &\emptycirc[0.5ex] \halfcirc[0.5ex] \fullcirc[0.5ex] & \emptycirc[0.5ex] \halfcirc[0.5ex] \fullcirc[0.5ex]\\
 
  Sinusoidal transformation of month/month of quarter  & \emptycirc[0.5ex] \halfcirc[0.5ex] \fullcirc[0.5ex]&\emptycirc[0.5ex] \halfcirc[0.5ex] \fullcirc[0.5ex] &\emptycirc[0.5ex] \halfcirc[0.5ex] \fullcirc[0.5ex] &\emptycirc[0.5ex] \halfcirc[0.5ex] \fullcirc[0.5ex] & \emptycirc[0.5ex] \halfcirc[0.5ex] \fullcirc[0.5ex]\\

\textbf{Life cycle} & & & & &\\
Normalized life cycle age & & &\emptycirc[0.5ex] \halfcirc[0.5ex] \fullcirc[0.5ex] &\emptycirc[0.5ex] \halfcirc[0.5ex] \fullcirc[0.5ex] & \emptycirc[0.5ex] \halfcirc[0.5ex] \fullcirc[0.5ex]\\

Volume-weighted life cycle age & & &\emptycirc[0.5ex] \halfcirc[0.5ex]  &\emptycirc[0.5ex] \halfcirc[0.5ex]  & \emptycirc[0.5ex] \halfcirc[0.5ex] \\

\textbf{Master} & & & & &\\
Cylinder  & & &\emptycirc[0.5ex] \halfcirc[0.5ex] \fullcirc[0.5ex] &\emptycirc[0.5ex] \halfcirc[0.5ex] \fullcirc[0.5ex] & \emptycirc[0.5ex] \halfcirc[0.5ex] \fullcirc[0.5ex]\\
Doors  & & &\emptycirc[0.5ex] \halfcirc[0.5ex] \fullcirc[0.5ex] &\emptycirc[0.5ex] \halfcirc[0.5ex] \fullcirc[0.5ex] & \emptycirc[0.5ex] \halfcirc[0.5ex] \fullcirc[0.5ex]\\
Relative performance  & & &\emptycirc[0.5ex] \halfcirc[0.5ex] \fullcirc[0.5ex] &\emptycirc[0.5ex] \halfcirc[0.5ex] \fullcirc[0.5ex] & \emptycirc[0.5ex] \halfcirc[0.5ex] \fullcirc[0.5ex]\\
Product type ID & & & & & \emptycirc[0.5ex] \halfcirc[0.5ex] \fullcirc[0.5ex]\\
Body type  & & & & & \emptycirc[0.5ex] \halfcirc[0.5ex] \fullcirc[0.5ex]\\
Fuel type  & & & & & \emptycirc[0.5ex] \halfcirc[0.5ex] \fullcirc[0.5ex]\\
Market cluster  & & \emptycirc[0.5ex] \halfcirc[0.5ex] \fullcirc[0.5ex]& \emptycirc[0.5ex] \halfcirc[0.5ex] \fullcirc[0.5ex]& \emptycirc[0.5ex] \halfcirc[0.5ex] \fullcirc[0.5ex]& \emptycirc[0.5ex] \halfcirc[0.5ex] \fullcirc[0.5ex]\\ 
Market maturity  & & \emptycirc[0.5ex] \halfcirc[0.5ex] \fullcirc[0.5ex]& \emptycirc[0.5ex] \halfcirc[0.5ex] \fullcirc[0.5ex]& \emptycirc[0.5ex] \halfcirc[0.5ex] \fullcirc[0.5ex]& \emptycirc[0.5ex] \halfcirc[0.5ex] \fullcirc[0.5ex]\\ 

\textbf{Internal} & & & & &\\
Dealer inventory  & \emptycirc[0.5ex] \halfcirc[0.5ex] & \emptycirc[0.5ex] \halfcirc[0.5ex] & \emptycirc[0.5ex] \halfcirc[0.5ex] & \emptycirc[0.5ex] \halfcirc[0.5ex] & \emptycirc[0.5ex] \halfcirc[0.5ex] \\ 
Dealer backorders  & \emptycirc[0.5ex] \halfcirc[0.5ex] & \emptycirc[0.5ex] \halfcirc[0.5ex] & \emptycirc[0.5ex] \halfcirc[0.5ex] & \emptycirc[0.5ex] \halfcirc[0.5ex] & \emptycirc[0.5ex] \halfcirc[0.5ex] \\ 
Fill rate ($\beta$-service level)  & \emptycirc[0.5ex] \halfcirc[0.5ex] & \emptycirc[0.5ex] \halfcirc[0.5ex] & \emptycirc[0.5ex] \halfcirc[0.5ex] & \emptycirc[0.5ex] \halfcirc[0.5ex] & \emptycirc[0.5ex] \halfcirc[0.5ex] \\ 
Sales targets  & \emptycirc[0.5ex] \halfcirc[0.5ex] \fullcirc[0.5ex] & \emptycirc[0.5ex] \halfcirc[0.5ex] \fullcirc[0.5ex]& \emptycirc[0.5ex] \halfcirc[0.5ex] \fullcirc[0.5ex]& \emptycirc[0.5ex] \halfcirc[0.5ex] \fullcirc[0.5ex] & \emptycirc[0.5ex] \halfcirc[0.5ex] \fullcirc[0.5ex] \\ 
Committed orders  & \emptycirc[0.5ex] \halfcirc[0.5ex] \fullcirc[0.5ex] & \emptycirc[0.5ex] \halfcirc[0.5ex] \fullcirc[0.5ex]& \emptycirc[0.5ex] \halfcirc[0.5ex] \fullcirc[0.5ex]& \emptycirc[0.5ex] \halfcirc[0.5ex] \fullcirc[0.5ex] & \emptycirc[0.5ex] \halfcirc[0.5ex] \fullcirc[0.5ex] \\ 
Net price  & & & \emptycirc[0.5ex] \halfcirc[0.5ex] \fullcirc[0.5ex]& \emptycirc[0.5ex] \halfcirc[0.5ex] \fullcirc[0.5ex] & \emptycirc[0.5ex] \halfcirc[0.5ex] \fullcirc[0.5ex] \\
Relative price to product cluster   & & & & \emptycirc[0.5ex] \halfcirc[0.5ex] \fullcirc[0.5ex] & \emptycirc[0.5ex] \halfcirc[0.5ex] \fullcirc[0.5ex] \\
Time since last price change   & & & & & \emptycirc[0.5ex] \halfcirc[0.5ex] \fullcirc[0.5ex] \\
\textbf{Online car configurator website traffic} & & & & &\\
Rolling mean (windows: 3)  & \emptycirc[0.5ex] \halfcirc[0.5ex]&\emptycirc[0.5ex] \halfcirc[0.5ex] & \emptycirc[0.5ex] \halfcirc[0.5ex] & \emptycirc[0.5ex] \halfcirc[0.5ex]  & \emptycirc[0.5ex] \halfcirc[0.5ex]  \\
Historic mean/standard deviation  & \emptycirc[0.5ex] \halfcirc[0.5ex] \fullcirc[0.5ex]&\emptycirc[0.5ex] \halfcirc[0.5ex] \fullcirc[0.5ex]& \emptycirc[0.5ex] \halfcirc[0.5ex] \fullcirc[0.5ex]& \emptycirc[0.5ex] \halfcirc[0.5ex] \fullcirc[0.5ex]  & \emptycirc[0.5ex] \halfcirc[0.5ex]  \fullcirc[0.5ex]\\
Volume-weighted website traffic  & \emptycirc[0.5ex] \halfcirc[0.5ex]&\emptycirc[0.5ex] \halfcirc[0.5ex] & \emptycirc[0.5ex] \halfcirc[0.5ex] & \emptycirc[0.5ex] \halfcirc[0.5ex]  & \emptycirc[0.5ex] \halfcirc[0.5ex]  \\
\textbf{Economy and financial markets} & & & & &\\
\textit{See Table \ref{tab: macro data}} & &\emptycirc[0.5ex] \halfcirc[0.5ex] \fullcirc[0.5ex]& \emptycirc[0.5ex] \halfcirc[0.5ex] \fullcirc[0.5ex]& \emptycirc[0.5ex] \halfcirc[0.5ex] \fullcirc[0.5ex]  & \emptycirc[0.5ex] \halfcirc[0.5ex]  \fullcirc[0.5ex]\\

      \bottomrule
    \end{tabular}
    }
    \caption{Data used as explanatory variables by hierarchy level.}
    \label{tab: internal data}
    \begin{minipage}{\textwidth}
        \footnotesize
        \textit{Notes.} 
        \emptycirc[0.5ex]: direct forecasting, \halfcirc[0.5ex]: hybrid forecasting, \fullcirc[0.5ex]: recursive forecasting \\
    \end{minipage}
\end{table}

\newpage
\subsection{Hierarchical time series}
\begin{figure}[H]
    \centering
    \includegraphics[scale=0.37]{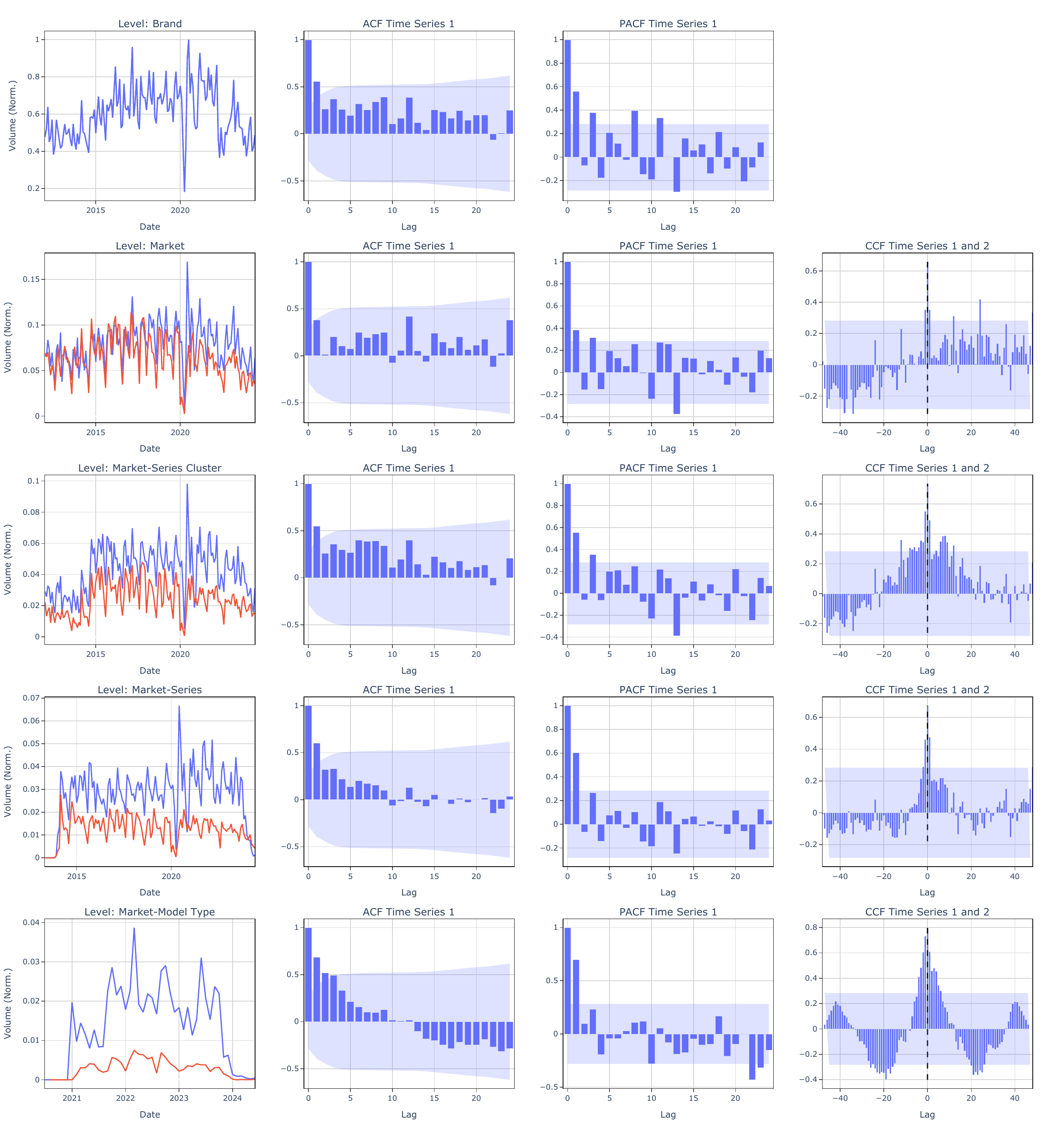}
    \caption{Plots of the time series, along with their ACF, PACF, and CCF.}\label{fig: ts autocorrelation}
    
    \vspace{0.5cm}
        \begin{minipage}{\textwidth}
            \footnotesize
            \textit{Notes.} The figure shows two example time series: series 1 (blue) and series 2 (red) across all hierarchy levels, differing only by market, not by product attributes. For each level, the autocorrelation (ACF), partial autocorrelation (PACF), and cross-correlation (CCF) are plotted. All data have been normalized for confidentiality. Visible anomalies reflect external shocks that affect forecasting accuracy, though their impact is largely mitigated in training by the length of the dataset.
        \end{minipage}

\end{figure}

\newpage
\subsection{Hyperparameter configurations}\label{apdx: hyperparameter}
\begin{table}[H]
\centering
\resizebox{0.95\textwidth}{!}{%
\footnotesize
\begin{tabular}{l l l l}
\toprule 
    Algorithm & Hyperparameter & Search space & Selected value \\ \hline 
    sNaive & Season length & & 12 \\ 
    MA & Window sizes & & $\{2,3,6,12\}$ \\ 
    ETS & Smoothing factor & $\{0.1, \dots, 1\}$ & 0.4\\ 
    ARIMA  & Season length & & 12 \\ 
    & \multicolumn{3}{l}{\textit{Other hyperparameters follow Nixtla’s AutoARIMA defaults.}} \\ 
    Prophet & Yearly seasonality &  & True\\ 
      & \multicolumn{3}{l}{\textit{Other hyperparameters follow the default settings of the Prophet model as implemented by Meta.}}\\ 
    DeepAR  & Input size && 12 \\
            & Objective && Quantile loss \\
            & Iterations && 1,000 \\
            & \# LSTM layers & $\{1,2,3\}$ & 2 \\
            & LSTM hidden size & $\{150, \dots,250\}$ & 200 \\ 
            & LSTM dropout rate & $\{0.01, \dots, 0.2\}$ & 0.1 \\
            & Decoder hidden layers & $\{1,2,3\}$ & 2 \\
            & Decoder hidden size & $\{150, \dots,250\}$ & 200 \\
            & Batch size & $\{32, \dots,128\}$& 40 \\
            & Learning rate & $\{0.00001 \dots, 0.1\}$ & 0.0001 \\ 
    NBEATSx & Input size &  & 12 \\
            & MLP units & & 512 \\
            & Iterations & & 1,000 \\
            & Loss function && Quantile loss \\
            & \# Blocks & $\{1,2,3 \}$& 1 \\
            & \# Harmonics & $\{2, \dots, 6\}$ & 5 \\
            & \# Polynominals & $\{2, \dots, 6\}$ & 2 \\
            & Learning rate & $\{0.00001 \dots, 0.1\}$ & 0.00609 \\
            & Activation function & $\{ \text{ReLU, Softplus, Tanh, Sigmoid}\}$ & Sigmoid \\ 
    LightGBM & Objective (point) && Regression \\
    & Objective (probabilistic) && Quantile \\
            & \# Estimators & \{50, \dots, 250\} & 153 \\
            & Max. leafs & \{10, \dots, 100\} & 77 \\
            & Max. depth & \{15, \dots, 100\} & 53 \\
            & Min. data in leaf & \{10, \dots, 200\} & 14 \\
            & Feature fraction & \{0.1, \dots, 1\} & 0.952 \\ 
            & Bagging fraction & \{0.1, \dots, 1\} & 0.178 \\
            & Learning rate & \{0.001, \dots, 0.1\} & 0.035 \\  
\bottomrule
\end{tabular}
}
\caption{Tuned hyperparameters selected for various forecasting models.}\label{Tab: hyperparameter}
\end{table}

\subsection{Forecast bias}\label{sec: forecast bias}

\subsection{Scaled pinball loss}\label{sec: pinball loss}

\end{document}